\documentclass[default,iicol]{sn-jnl}

\usepackage{graphicx}%
\usepackage{multirow}%
\usepackage{amsmath,amssymb,amsfonts}%
\usepackage{amsthm}%
\usepackage{mathrsfs}%
\usepackage[title]{appendix}%
\usepackage{xcolor}%
\usepackage{textcomp}%
\usepackage{manyfoot}%
\usepackage{booktabs}%
\usepackage{algorithm}%
\usepackage{algorithmicx}%
\usepackage{algpseudocode}%
\usepackage{listings}%
\usepackage{natbib}

\usepackage{xcolor}

\usepackage[export]{adjustbox}

\theoremstyle{thmstyleone}%
\theoremstyle{thmstyletwo}%

\theoremstyle{thmstylethree}%

\begin{document}

\title{Multi-level Relation Learning for Cross-domain Few-shot Hyperspectral Image Classification}


\author[1,2,3,4]{\fnm{Chun} \sur{Liu}}\email{liuchun@henu.edu.cn}

\author[1]{\fnm{Longwei} \sur{Yang}}\email{104754211314@henu.edu.cn}

\author[1,2,3,4]{\fnm{Zheng} \sur{Li}}\email{lizheng@henu.edu.cn}
\equalcont{These authors contributed equally to this work.}

\author[1,2,3,4]{\fnm{Wei} \sur{Yang}}\email{weiyang@henu.edu.cn}
\equalcont{These authors contributed equally to this work.}

\author[4,5]{\fnm{Zhigang} \sur{Han}}\email{zghan@henu.edu.cn}
\equalcont{These authors contributed equally to this work.}

\author*[4,5]{\fnm{Jianzhong} \sur{Guo}}\email{guojianzhong@vip.henu.edu.cn}
\equalcont{These authors contributed equally to this work.}

\author[6]{\fnm{Junyong} \sur{Yu}}\email{yujunyong@163.com}
\equalcont{These authors contributed equally to this work.}

\affil[1]{\orgdiv{School of Computer and Information Engineering}, \orgname{Henan University}, \orgaddress{\city{Zhengzhou}, \postcode{450046}, \state{Henan}, \country{China}}}

\affil[2]{\orgdiv{Henan Key Laboratory of Big Data Analysis and Processing}, \orgname{Henan University}, \orgaddress{\city{Zhengzhou}, \postcode{450046}, \state{Henan}, \country{China}}}

\affil[3]{\orgdiv{Henan Engineering Laboratory of Spatial Information Processing}, \orgname{Henan University}, \orgaddress{\city{Zhengzhou}, \postcode{450046}, \state{Henan}, \country{China}}}

\affil[4]{\orgdiv{Henan Industrial Technology Academy of Spatio-Temporal Big Data}, \orgname{Henan University}, \orgaddress{\city{Zhengzhou}, \postcode{450046}, \state{Henan}, \country{China}}}

\affil*[5]{\orgdiv{College of Geography and Environmental Science}, \orgname{Henan University}, \orgaddress{\city{Zhengzhou}, \postcode{450046}, \state{Henan}, \country{China}}}

\affil[6]{\orgname{Henan DeFan High-tech Software Co., Ltd}, \orgaddress{\city{Zhengzhou}, \postcode{450046}, \state{Henan}, \country{China}}}


\abstract{Cross-domain few-shot hyperspectral image classification focuses  on learning prior knowledge from  a large number of labeled samples from source domains and then transferring the knowledge to the tasks which contain few labeled samples in target domains.  Following the metric-based manner, many current methods first extract the features of the query and support samples, and then directly predict the classes of query samples according to their distance to the support samples or prototypes. The relations between  samples have not been fully explored and utilized. Different from current works, this paper proposes to learn sample relations on different levels and take them into the model learning process, to improve the cross-domain few-shot hyperspectral image classification. Building on  current method of "Deep Cross-Domain Few-Shot Learning for Hyperspectral Image Classification" which adopts a domain discriminator to deal with domain-level distribution difference, the proposed method applies contrastive learning to learn the class-level sample relations to obtain more discriminable sample features. In addition, it adopts a transformer based cross-attention learning module to learn the set-level sample relations and  acquire the attention from query samples to support samples. Our experimental results have demonstrated the contribution of the multi-level relation learning mechanism for few-shot hyperspectral image classification when compared with the state of the art methods. All the codes are available at github https://github.com/HENULWY/STBDIP.}




\maketitle

\section{Introduction}

Hyperspectral remote sensors obtain radiation information of ground objects on dozens or even hundreds of spectral bands. While imaging the ground objects, they can obtain  continuous spectral information at each pixel, resulting in that  hyperspectral images contain both spatial  and spectral information.  Because spectral information can reflect the unique physical characteristics of ground objects,  hyperspectral images can capture more subtle ground differences  compared with traditional visual images. This  makes that  hyperspectral images are widely used in land cover analysis, environmental monitoring, geological survey, military reconnaissance and other fields.

As one of the fundamental tasks for hyperspectral image applications, hyperspectral image classification is to classify each pixel of hyperspectral images and recognize the captured objects, which has received extensive attention in the past decades \citep{1,2}.  Early hyperspectral image classification methods mainly focused on the spectral information, and applied the classifiers such as support vector machine (SVM) \citep{3} and logical regression \citep{4} directly to spectral vectors. Due to the data redundancy of high-dimensional spectral information, researchers have also applied the techniques like principal component analysis \citep{5} and  linear discriminant analysis \citep{6} to reduce the dimensionality of  spectral vectors. In light of  that  spatial information are helpful to improve the  classification accuracy,  current  methods tend to  integrate spatial and spectral information for hyperspectral image classification \citep{7,8}.

In recent years, with great success in the fields of  image classification, natural language processing, speech recognition, etc., deep learning  has been widely used in hyperspectral image classification \citep{1,9,10}  and achieved the state of art  performance. Through a multi-layer neural network which transforms the initial low level features  into high level features, deep learning can learn more abstract and distinguishable spatial and spectral features from hyperspectral images.   Considering the  different ways of extracting and fusing spectral and spatial features,  a series of  hyperspectral image classification methods have been presented by using different deep learning models such as automatic encoders \citep{11,12}, convolutional neural networks \citep {9,13,14,15,16} ,  recurrent neural network \citep{17,18,19} and  graph convolution network \citep{20,21,22}.

Due to the large number of model parameters,   deep learning  methods  usually  require a large number of labeled samples for training. This brings  challenge to the application of these methods because  it  is  often expensive and time-consuming to mark enough samples. In contrast,  humans can identify new samples by only learning from a small number of labeled samples. In light of that, few-shot learning methods have aroused wide interest recently, which aim at acquiring prior knowledge from  a large number of labeled samples and then transfer the knowledge to target tasks which contain few labeled samples. To address few-shot hyperspectral image classification, many  methods have been proposed based on the works on few-shot learning  \citep{2}. For example, in the metric-based manner, Tang et al. \citep{23} introduced the prototype network method for  hyperspectral image classification; Ma et al. \citep{24} applied the relational network  to achieve hyperspectral image classification under the few-shot setting.

When addressing few-shot hyperspectral image classification, it is often  assumed that the samples from which the  prior knowledge is learned and the samples of the  target tasks come from the same domain.  But it is  difficult to guarantee that  in actual application scenarios. For example,  hyperspectral images are often affected by the atmospheric environment and the  sensors. Different hyperspectral image datasets are often from different domains and have different data distributions. Due to the  distribution difference between different domains, there will be a decline in performance when applying the model learned from the datasets in one domain to the target tasks in another domain. Therefore, to further address the cross-domain few-shot problem, some methods have also been  proposed for hyperspectral image classification by enhancing the  cross-domain  generalization capability. For example, Li et al. \citep{25} have followed the generate adversarial network to design and train the model, with the aim at making that the model can extract domain-independent features. At the same time, Bai et al. \citep{26} have devised  a feature-wise transformation module to change the  distribution of extracted features and make the feature representations  more generalized.

However,  by following the metric-based few-shot learning manner, many current methods  for  few-shot hyperspectral image classification  first extract the sample features and then directly predict the classes of query samples  according to their distance to the support samples or prototypes in embedding space. The relations between  samples have not been fully explored. The predictions depend heavily  on the  quality of the extracted sample  features.  Although Li et al. \citep{25} and Zhang et al.  \citep{48} have noticed the domain-level sample relations, i.e., the distribution difference between different domain, the sample relations  within domain need to be further excavated. And more benefits will be there when  these relations are learned and  used to constrain the model learning process. For example, if we measure the class-level sample relations and make that the samples from the same classes are closer and from different classes are father in the embedding space, more discriminative features will be generated, which contributes to  downstream distance based prediction. At the same time, taking the set-level sample relations into consideration, if we let the samples from query set take a look at the samples from support set before distance measurement to learn about which support sample is more similar and highlight the similar parts accordingly,  the performance of distance based prediction will be also improved. 

Therefore, this paper proposes to learn sample relations from different levels and take them into the model learning process, to improve the cross-domain few-shot hyperspectral image classification. Building on the method proposed by Li et al. \citep{25} which adopts the domain discriminator to deal with the domain-level distribution difference, the proposed method applies contrastive learning \citep{54} to learn the class-level sample relations, with the aim at improving inter-class discriminability of sample features. In addition, it adopts a transformer based cross-attention learning module to learn the set-level sample relations and accordingly acquire the attention from query samples to support samples. Our experimental results have demonstrated the contribution of the multi-level relation learning mechanism for few-shot  image classification when compared with the state of the art methods. 

The main contributions of this work are as follows.
\begin{enumerate}

\item We take into account the sample relation learning for few-shot  hyperspectral images classification, and build a multi-level relation learning framework to improve the feature discriminability.

\item We propose the contrastive learning at  class-level, and the cross-attention learning at set-level, to make the samples from the same clases closer and the samples from different classes father.

\item Extensive experiments have been done and the results demonstrate the capability of the multi-level relation learning framework and their contributions to few-shot  hyperspectral images classification.
\end{enumerate}

The remainder of this paper is structured as follows. Section 2 introduces the related works. Section 3 details the proposed method. Section 4 describes the datasets, the design and the results of the experiments. Finally, the conclusions are in section 5.

\section{Related works}

The work in this paper involves few-shot learning, deep learning based hyperspectral image classification, and few-shot hyperspectral images classification. In this section, we give a brief introduction about these related works. 

\subsection{Few-shot Learning}

Few-shot leaning aims to solve new tasks with few labeled data based on the knowledge obtained from previous experiences. It  utilizes the available datasets which contain  a large number of labeled samples, and learns a model which has good generalization ability and can quickly adapt to new tasks. The means commonly used  is the  episode-based training method.

Episode-based training method believes that the testing environment of machine learning should be consistent with the training environment to ensure the generalization ability. Therefore, it constructs many tasks similar to the target tasks from the available datasets, and takes the constructed tasks as samples to train the model. Specifically, both the target tasks and the construncted tasks  include two sets of samples: support set and query set. The  support set contains the  few available labeled samples. If there are $C$ classes of samples  and $K$ samples per-class in the support set, the tasks are  called $C$-way $K$-shot tasks. In the meanwhile, the query set contains the samples whose labels are to be predicted according to their matching with these samples in the support set. Using the target tasks and the constructed  tasks, the episode-based training process is divided into two stages: meta-train stage and meta-test stage. Meta-train stage is to train the model by using the constructed  tasks, while meta-test stage is to further fine-tune the trained model and test its performance  by using the  samples of the target tasks.

At present,   few-shot learning has attracted much attention from researchers in different fields, including image classification, object detection, action recognition, semantic segmentation, etc.  By focusing on the works of image classification in computer vision field, current few-shot  learning  methods can be roughly divided into three categories: \textit{metric-based methods}, \textit{optimization-based methods} and\textit{ graph-based methods}. \textit{Metric-based methods}  use the base class samples to obtain a distance function that can measure the similarity between samples but also  has good generalization ability. This distance function can migrate well to the new classes of samples. Through this distance function,  the classes of unlabeled samples in the target tasks can be inferred based on their similarity with  few  labeled samples\citep{27,28,29,30,31,32,33}.  Different from metric-based methods, \textit{ optimization based methods} use the constructed tasks to train an optimized model in order to obtain an initialization of  model parameters. With the parameter initialization, the few labeled samples of the target tasks can be used to adjust the model and make it  adapted to the target tasks \citep{34,35}. In the meanwhile, in light of that both  metric-based methods and optimization-based methods ignore the relations between samples, the \textit{graph based methods} use graph to establish  the relations between samples, and then the  labels of labeled samples migrate to unlabeled samples through graph convolution networks or label propagation algorithm \citep{36,37,38,39,40}. Because it is easy to understand, most few-shot  methods are following the metric-based way. In this paper, we also adopt the metric-based way to design the few-shot methods for hyperspectral image classification.

\subsection{Deep Learning based Hyperspectral Image Classification}

Due to the powerful feature extraction ability, deep learning has obtained great success in many applications and is also widely used for hyperspectral image classification. Many classification methods have been proposed for hyperspectral images based on different deep learning models such as automatic encoder, convolutional neural network, recurrent  neural network and graph convolution network. Their purposes are to use these models to extract better spatial and spectral features from hyperspectral images. We give a brief introduction about these works in this subsection, and for more information, please refer to  the works of Li et al. \citep{1} and Jia et al. \citep{2}.

Automatic encoder (AE)  is composed of an encoder and a decoder. By using the encoder to encode the input data and the decoder to reconstruct the input data, it can generate the feature representations of input data. Thus, the AE based methods  adopt AE to obtain the spatial and spectral feature representations, or spectral-spatial feature representations  of hyperspectral images, especially when there are none labeled samples available \citep{11,12}. Using a set of filters to scan images and generate specified features, convolutional neural network (CNN)  has shown a strong feature extraction ability with less parameters. The CNN based methods  mainly use different convolution operators (i.e., 2D, 1D, 3D operator) \citep{13,14,15,16} or different convolution networks (i.e., residual network or dense network) \citep{15,41} to extract spatial and spectral features from hyperspectral images. Compared with CNN, recurrent neural network (RNN)  is designed for extracting features from  sequence data. For hyperspectral images, each hyperspectral image can be regarded as a sequence containing many pixels, and each pixel is also a sequence containing many band data. Therefore, many researchers have also used the typical RNN model such as  LSTM and GRU to extract  spatial or spectral features of hyperspectral images \citep{17,18,19}. Graph convolution network (GCN) is a model for processing graph  data based on deep learning. In recent years,  researchers have also used pixels or hyperpixels of hyperspectral images as vertexes to construct graph, and then used graph convolution network to extract the features of pixels or hyperpixels, which contained the contextual information from other  pixels or hyperpixels \citep{20,21,22} .

\subsection{Few-shot Hyperspectral Image Classification}

Because it is difficult to label enough samples for each class of ground objects captured by  hyperspectral images, many research works have been done in recent years to address  few-shot  hyperspectral image classificaiton. These works are mainly based on the typical few-shot learning models of  twin network \citep{27}, prototype network \citep{29},  relational network \citep{30} and graph convolutional network \citep{22,63}.

With two neural networks sharing parameters, twin network receives a pair of samples each time, obtains their embedded representations, and finally calculates the similarity between them based on a simple distance function. By setting the  output to be 1 when the pair of samples is from the same classes and 0 when they are from different classes, a model  is trained which has good generalization ability and can be used for addressing few-shot problem. Based on the twin network, researchers have used 2D CNN\citep{42}, or combined 1D CNN with 2D CNN \citep{43}, or directly  used 3D CNN\citep{44} as the backbone network of twin network to achieve hyperspectral images classification.  Prototype network believes that each class of samples possesses a representing prototype in the embedding space. It predicts the sample classes by calculating their distance to different class prototypes. Based on the prototype network architecture, many methods have  been designed for hyperspectral images classification \citep{23,45}. For example, considering the characteristics of hyperspectral images, Tang et al. has presented the local pattern coding \citep{23} to extract features for hyperspectral image classification; Sun et al. has proposed the Earth Mover's distance to replace the Euclidean distance to further improve the classification performance \citep{45}. Based on prototype network, relational network adds a relational module  to learn a suitable function for measuring the distance between  samples and class prototypes instead of  the fixed  functions. Similar with the works based on Twin network,  the research works based on relational networks also use different embedding methods such as 1D and 2D CNN \citep{24}, or 3D CNN \citep{46,47} as feature extraction modules to achieve the classification of hyperspectral images. Different from above works, the GCN based methods for hyperspectral image classification construct the graph by using the samples in the tasks, and then execute the graph convolutions to obtain the sample features and infer the labels of query samples \citep{22, 62}.

To further address cross-domain few-shot  problem,  Li et al. \citep{25} have augmented the prototype network with a discriminator to enable that the model can extract domain-independent features. At the same time, while proposing a 3D local channel attention residual network for obtaining the spatial-spectral features of hyperspectral images and a subspace classifier for classification, Bai et al. \citep{26} have devised  a feature-wise transformation module to change the  distribution of extracted features and make the feature representations  more generalized under the cross-domain setting.   In the meanwhile, Xi et al. \citep{61}  devised a spectral prior-based refinement module and a lightweight cross-scale convolutional network  to capture feature repretations from hyperspectral images. Additionally, they used the Mahalanobis distance measurement rather than an extra classifier to infer the class labels.  When adopting the intra-domain distribution extraction block  and cross-domain similarity aware block to  aggregate the intra-domain  relationships and capture the inter-domain feature  similarities, Zhang et al.  \citep{48} proposed to mitigate the impact of domain shift on few-shot hyperspectral images classification through graph alignments.  Different from these works,  the proposed method in this paper aims to explore the different kinds of sample relations under the ''feature extractor+distance metric" architecture, and study how to utlize them to enhance the model learning process for cross-domain few-shot hyperspectral image classification. 

\section{The proposed method}

In this section, we state the problem of cross-domain few-shot  hyperspectral image classification and detail the method proposed.

\subsection{Problem statement}

Hyperspectral image classification is to predict the classes of each pixel of the images. That is,  the image pixels  are the samples to be classified. Currently, in order to utilize the spatial  information for classification, the fixed-size patches which center around each pixel instead of the pixels themselves are  taken as the samples for classification. Because it is costly to mark  enough pixels for training a model which maps  the pixels to their classes, few-shot hyperspectral images classification problem is about whether we can predict  pixel classes by only marking few of them, for example, marking one or three pixels per class. Following few-shot learning, these marked few pixels form the support set of the few-shot classification task, and the remaining unlabeled pixels constitute the query set. As mentioned earlier,  if there are $C$ classes and  $K$ labeled pixels  per class, this few-shot classification task is  called $C-way$ $ K-shot $ task.

To resolve the few-shot classification task, it needs a matching model $f(.)$ which can predict the classes of query samples according to their matching with these support samples. To learn the model $f(.)$, few-shot learning resorts to a source dataset $D_{s}$ which consists of a large number of labeled samples. $D_{s}$  has   more classes and  more labeled samples per class. And to ensure the generalization of learned model, it mimics the target task and constructs a large number of tasks from $D_{s}$. That is, each constructed task is also composed by a support set and a query set,  both of which have the same number of samples as these of target task. The difference is that, all the samples in the constructed tasks are labeled.  With these constructed tasks, it uses the samples from one task  each time to train and update the model $f(.)$. For the cross-domain few-shot classification problem, the samples of  target task and that of  $D_{s}$ are from different domain with different data distribution.

Once trained on $D_{s}$, the matching model $f(.)$ will be tested on target tasks. Generally, the model  $f(.)$ will be tested by a large number of tasks which are constructed from a target dataset $D_t$ and the average performance is adopted. There is often none intersection between $D_{s}$ and $D_t$, i.e., $D_{s} \cap D_t $ = $\emptyset$. 

\subsection{Model Architecture}

\begin{figure*}[h]
	\centering
	\includegraphics[width=16cm]{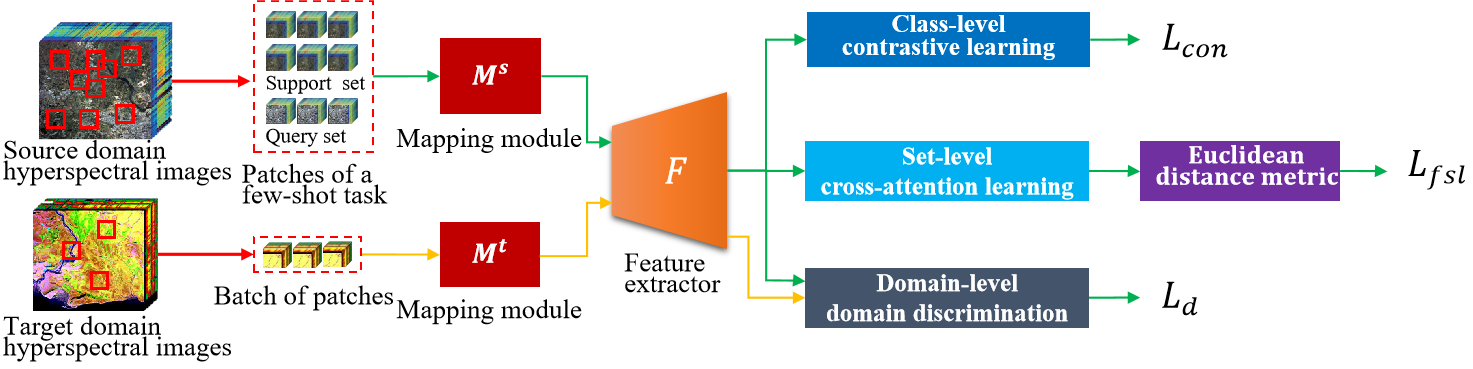}
	\caption{The model architecture  of proposed method.}\label{structure3}
\end{figure*}

Building on the work of Li et al. \citep{25}, the model architecture of the proposed method is shown in Fig \ref{structure3}. The input include a  $C-way$ $ K-shot $ few-shot task which consists of a set of support and query patches from one domain,  and  a batch of patches from the other domain. With these input, there are two \textit{mapping modules} first to  transform the patches from source domains and target domains into new patches with the same number of bands. Then, the features of these new patches are extracted by \textit{feature extractor}. Following the feature extractor, there are three modules which deal with the sample relations from different levels.

\begin{enumerate}
	\item	Taking the features of the patches within the  few-shot classification task as input, the \textit{contrastive learning module} measures the class-level sample relations and generates the  contrastive loss $L_{con}$.
	
	\item Focusing on the set-level sample relations between the query set and support set within the few-shot classification task, the \textit{cross-attention learning module} strives to  acquire the attention from query patches to the support patches and update the features of query patches accordingly. Given the support features and the updated query features, the Euclidean  distance metric is then applied to infer the labels of query patches and the few-shot classification loss $L_{fsl}$ is produced. 
	
	\item Taking the features of all these patches from source and target domains together as input,  the \textit{domain discriminator}, which addresses the domain-level sample relations,  discriminates the features of one domain from these of another domain.  In this process, the domain discrimination loss $L_{d}$ will be generated.
\end{enumerate}

According to the domain of the input few-shot  tasks, i.e.,  from source domains or target domains,  there are two computing streams which execute alternately in training stage, i.e., the stream for  source domain and the stream for target domain.  And there are four steps to be executed for each computing stream: mapping and feature extracting, class-level contrastive learning, set-level cross-attention based few-shot classification, and domain-level domain discrimination. We detail them as follows.

\subsubsection{Mapping and Feature Extracting}
As stated before, mapping and feature extracting are to map the patches into embedding space and obtain their deep features. They are executed by the mapping modules and the feature extractor shown in Fig \ref{structure3}.  

The mapping modules are implemented with a Conv2D layer with $1 \times 1 $ kernel and a BatchNorm layer. Given a patch with the size of $9 \times 9 \times d_1$, the mapping modules  transform it into a new patch with the size of $9 \times 9 \times d_2$.   The feature extractor is implemented by  3D CNN, which consists of  two residual blocks and a convolutional layer. The implementation of these residual blocks and convolutional layer  are shown in Table \ref{extraction-module}. 

\begin{table}
	\caption{The architecture of feature extractor}\label{extraction-module}
	\setlength{\tabcolsep}{1mm}{
		\begin{tabular}{cc|c}
			\toprule
			\multicolumn{2}{c|}{ Feature Extractor Architecture} & SIZE \\ \midrule
			\multicolumn{2}{c|}{INPUT} & 9$\times$9$\times d_2$ \\ \midrule
			\multirow{7}{*}{Residual block1} & Conv3d & 3$\times$3$\times$3(8) \\
			& BatchNorm3d + Relu & N/A \\
			& Conv3d & 3$\times$3$\times$3(8) \\
			& BatchNorm3d + Relu & N/A \\         
			& Conv3d & 3$\times$3$\times$3(8) \\
			& BatchNorm3d + Relu & N/A \\
			& MaxPool3d & 4$\times$2$\times$2 \\ \midrule
			\multirow{7}{*}{Residual block2} & Conv3d & 3$\times$3$\times$3(16) \\
			& BatchNorm3d + Relu & N/A \\
			& Conv3d & 3$\times$3$\times$3(16) \\
			& BatchNorm3d + Relu & N/A \\         
			& Conv3d & 3$\times$3$\times$3(16) \\
			& BatchNorm3d + Relu & N/A \\
			& MaxPool3d & 4$\times$2$\times$2 \\ \midrule
			\multirow{1}{*}{Convolutional layer} & Conv3d & 3$\times$3$\times$3(32) \\ \midrule
			\multicolumn{2}{c|}{OUTPUT} & 1$\times$160 \\ \bottomrule
		\end{tabular}
	}
\end{table}

\subsubsection{Class-level Contrastive  Learning}
Class-level contrastive  learning is to measure the contrastive relations between the patches of the input few-shot task in the embedding space. By following the supervised contrastive learning \citep{54}, the contrastive loss $L_{con}$ will be produced.

Taking a patch as anchor, the contrastive loss tends to make it close to the patches from the same class in the embedding space, and apart from the patches from other classes. Given the patches of the input task whose normalized  features are $z$, the  contrastive loss $L_{con}$ is defined by Eq.(1).

\begin{equation}
	\resizebox{\columnwidth}{!}{$
		\textit{\L}_{con}=-\sum_{i\in I}\frac{1}{|P(i)|}\sum_{p\in P(i)} log\frac{exp(z_{i} \bullet z_{p}/\tau)}{\sum_{a\in I \backslash i}exp(z_{i}\bullet z_a/\tau)}
	$}
\end{equation}

In above definition, $i\in I=|z|$ is the index of an arbitrary patch of the input task, and $z_i$ is the normalized feature of the $i^{th}$ patch. $P(i)=\{p\in I\backslash i\}$ is the set of patch indices where the patches  are from the same class as that of the $i^{th}$ patch. And, $\tau \in R^+$ is a temperature parameter controlling the concentration of contrastive loss on hard samples. Because the  features are normalized, the dot production of $z_{i} \bullet z_{p}$ means the cosine similarity between $z_{i}$ and $z_{p}$. Then the  contrastive loss $L_{con}$ will constrain the feature extractor to enlarge the similarity between the features from the same classes, and lower the similarity between the features from different classes. In this way, it  promotes the  discriminability of the features  extracted. 

\subsubsection{Set-level Cross-attention based Few-shot Classification}
\begin{figure*}[h]
	\centering
	\includegraphics[width=12cm]{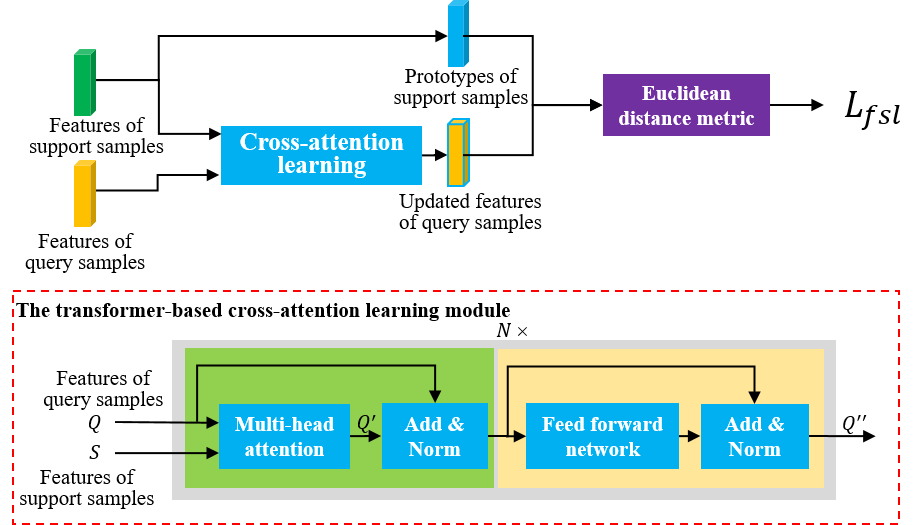}
	\caption{\textbf{The cross-attention based  few-shot classification}. The upper part shows the process of the few-shot classification. The bottom part is the architecture of the cross-attention learning module which is implemented by following transformer. While  the features of query samples are defined as $Q$, the features of support samples are denoted by $S$. And $Q'$ represents the query sample features  updated by the attention layer in Transformer,  $Q''$ represents the query set sample features produced by a Transformer layer which includes attention and feedforward layers. }\label{affinitylearning}
\end{figure*}

Cross-attention based few-shot classification is to learn the  relations between the support and query patches,  infer the labels of query patches and produce the few-shot learning loss $L_{fsl}$. As shown in Fig \ref{affinitylearning}, with the cross-attention learning module,  the attention from query patches to support patches are obtained,  and then the features of query patches are updated accordingly. At the same time, assuming the feature of the $j^{th}$ patch of $i^{th}$ class is $z_{ij}$, the prototypes of each class of the support patches are generated as  Eq.(2), where $k$ is the number of support patches of  $i^{th}$ class. With the updated query features and the prototypes of each support class, the Euclidean distance between the query features and the prototypes are  calculated, and  the  few-shot learning loss will be generated accordingly by Eq.(3). The $x_i$ in Eq.(3) represents $i^{th}$ query patch, whose label  $y_i=h$. $N$ is the number of query patches. And  $p(y_i=h|x_i)$ means the probability that the model accurately predicts the label of $i^{th}$ query patch.  Meanwhile,   $d(z_i, p_h)$ denotes the  Euclidean distance between the feature $z_i$ of $i^{th}$ query patch and the prototype $p_h$ of $h^{th}$ support class. 

\begin{equation}
	p_i= \frac{1}{k} \sum^k_{j=1} z_{ij}
\end{equation}

\begin{equation}
	\begin{aligned}
		\textit{\L}_{fsl}=- \frac{1}{N} \sum^{N}_{i=1} log p(y_i =h | x_i) \\
		where \quad p(y_i =h | x_i) = \frac{exp( d(z_i, p_h))}{\sum^k_{j=1}exp(d(z_i, p_j))}
	\end{aligned}
\end{equation}

We follow transformer \citep{55} to acquire the attention from query patches to support patches. That is,  the scaled dot-product attention is adopted for implementing the cross-attention learning module,  as shown in the bottom of Fig \ref{affinitylearning}.  It is composed of several identical layers (i.e., 2 layers in proposed method). Each layer has two sub-layers. The first is a multi-head attention layer, and the second is a  feed-forward network. In each sub-layer, the residual connection (i.e., the Add operation) is used, and followed is the layer normalization. 

Given a set of query $Q$, attention is a function that  maps each query and a set of key-value pairs $(K, V)$ to an output, where the output is the weighted sum of all the values in $V$ \citep{55}.  As shown  in Eq.(4) , the weights on each value are calculated by computing the dot-product of each key  to the query, dividing them by the scaling factor $\sqrt{d_k}$ where $d_k$ is the dimension of the key vector, and finally applying the softmax function. To allow the model to jointly attend to information from different  subspaces, multi-head attention shown in Eq.(5) applies multiple attention functions shown in Eq.(4) (e.g., 8 heads in our method)  and then concatenates the outputs together. The $W^Q_i$, $W^K_i$ and $W^V_i$ are the learnable matrices that project input $Q$, $K$ and $V$ into different subspaces. After using a  projecting matrix $W^o$, the concatenated output is mapped into a new vector with desired dimension. 

\begin{equation}
	Attention(Q, K, V)= softmax(\frac{QK^T}{\sqrt{d_k}}) V
\end{equation}

\begin{equation}
	\resizebox{\columnwidth}{!}{$
	\begin{split}
		MultiHeadAtt(Q, K, V)= Concat(head_1, \dots, head_h) W^{o} \\
		where \quad  head_i = Attention(QW^Q_i, KW^K_i, VW^V_i)
	\end{split}
	$}
\end{equation}

Given a set of features of query patches $Q$ and a set of features of support patches  $S$ in a few-shot hyperspectral image classification task, we acquire the attention of query patches to support patches and use the attention to update the features of query patches by allowing $K$ and $V$ in Eq.(5) to be equal to $S$ as shown in Eq.(6), i.e., $K=V=S$. 

\begin{equation}
\resizebox{\columnwidth}{!}{$
	\begin{split}
		MultiHeadAtt(Q, S)= Concat(head_1, \dots, head_h) W^{o} \\
		where \quad  head_i = Attention(QW^Q_i, SW^K_i, SW^V_i)
	\end{split}
$}
\end{equation}

\subsubsection{Domain-level Domain Discrimination} 
Domain-level domain discrimination is to discriminate the features of one domain   from these of another domain, and accordingly generate the discriminator loss $L_d$.   Its purpose is to make that the feature extractor can produce more generalized features.  Li et al. have proposed a conditional domain discriminator for this purpose by following the  conditional domain adversarial network (CDAN) \citep{25,56}. Their implementations for the conditional domain discriminator  are kept in our work. For more information about it, please refer to the work of Li et al.  \citep{25} .

\subsection{The Overal Loss Function}

After above four steps of  computing in one iteration, the follow total loss is used to update the model, where $\lambda_i$($i=1,2,3$) are the weights assigned to these three kinds of losses. The whole training  process of the proposed model is shown in Algorithm 1.

\begin{equation}	
	L = \lambda_1 L_{con} + \lambda_2 L_{fsl} + \lambda_3 L_{d}
\end{equation}

\begin{algorithm}

	\caption{Multi-level Relation Learning Training Process}\label{algo1}
	\begin{algorithmic}[1]
		\Require {
			The feature extrator $f$ \newline
			The source sample dataset $D_s$ \newline
			The target smaple dataset $D_t$ \newline
			The support set $S_s$ and query set $Q_s$ from $D_s$  \newline
			The support set $S_t$ and query set $Q_t$ from $D_t$ }
		\Ensure The loss of model
		\While {epoch $<$ 10000}
		\If{epoch$ \% 2 == 0$}
		\State Train with $S_s$, $Q_s$ and $D_t$
		\State	Calculate $L_{con}$ of an episode by Eq.(1)
		\State	Update $Q_s$ using $S_s$ by Eq.(6)
		\State	Calculate $L_{fsl}$ of an episode by Eq.(3)
		\State	Calculate $L_{d}$ of an episode
		\State	Calculate $\lambda_1 L_{con} + \lambda_2 L_{fsl} + \lambda_3 L_{d}$	
		\Else
		\State {Train with the same process as above as using $S_t$, $Q_t$ and $D_s$}
		\EndIf
		\EndWhile
	\end{algorithmic}

\end{algorithm}

\subsection{The Testing}
\begin{figure*}[htbp]
	\centering
	\includegraphics[width=13cm,height=2.5cm]{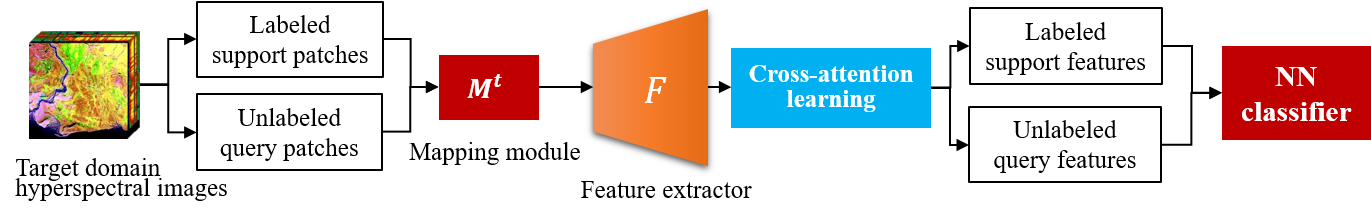}
	\caption{The testing process of the proposed method.  }\label{testingprocess}
\end{figure*}

Once trained, the  mapping module for target domain $M^t$, the feature extractor and the cross-attention learning module shown in Fig \ref{structure3} are kept in the testing stage. As shown in Fig \ref{testingprocess},  the few-shot tasks which consist of the support patches (i.e., labeled patches) and the query patches (i.e., unlabeled patches) are first constructed from the target domain hyperspectral images. And then, the features of the support and query patches are extracted through the mapping module and the  feature extractor.  After that, the features of query patches are updated through the cross-attention learning module. Finally, given the features of these support and query patches, the classes of the query patches will be predicted by a nearest neighbor(NN) classifier.

\section{Experiments}

To validate the proposed method, several experiments have been done in our work. In this section, we describe the experimental setup and  results.

\subsection{Experimental Setup}

\subsubsection{Dataset}

In order to fairly evaluate the performance of the proposed method, the datasets which have been widely used in related works are selected for training and testing in our work. Specifically, we use the Chikusei dataset as the source domain dataset, with the Indian Pines, Pavia University and Salinas datasets as the target domain datasets. The data can be obtained in \citep{26}. We give a brief introduction about these datasets. 

\begin{figure}[htbp]
	\centering
	\includegraphics[width=7cm]{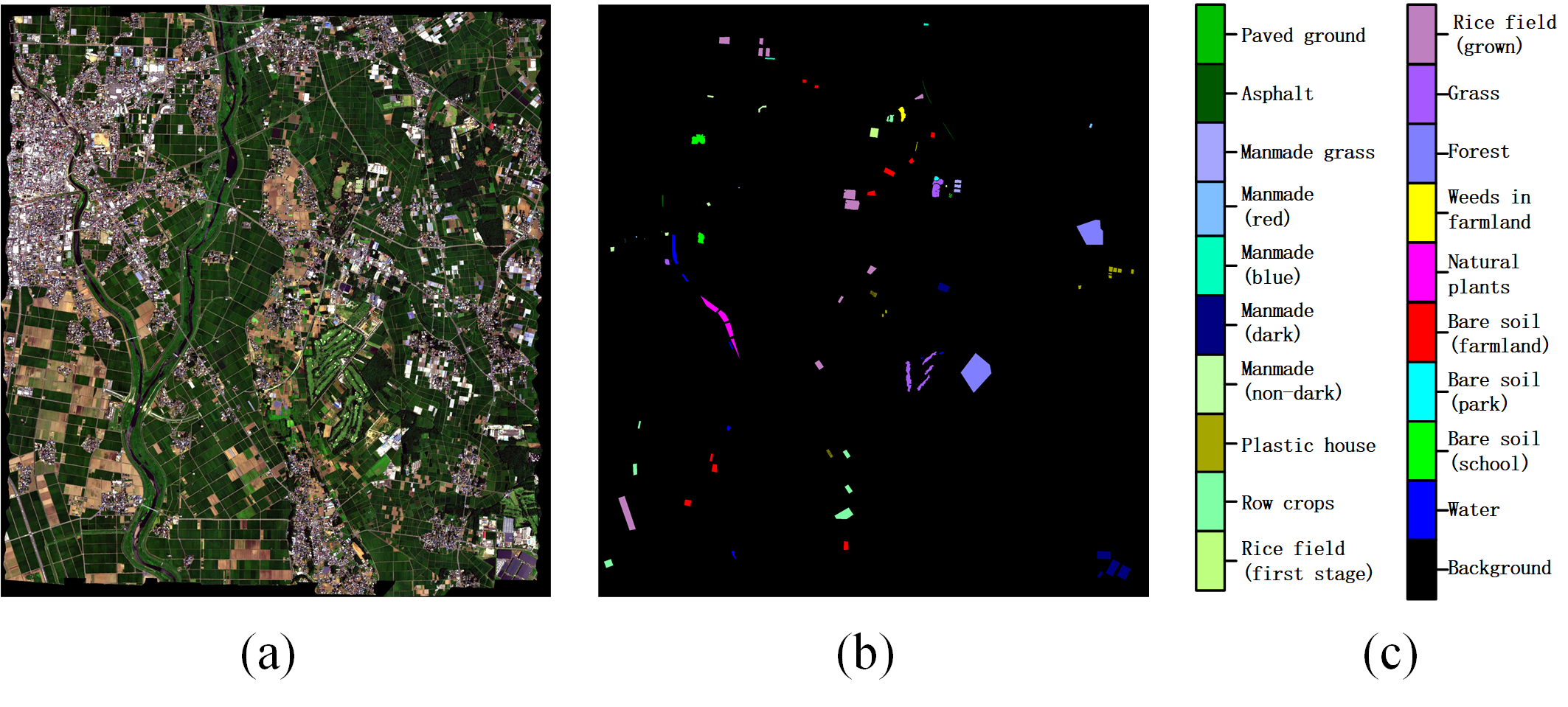}
	\caption{  The color map, label map and label color of Chikusei dataset}\label{fig1}
\end{figure}

\textbf{Chikusei}: It was taken by Hyperspec-VNIR-C sensor in Chikusei, Japan. The data contains 128 bands with the wavelength ranging from $343nm$ to $1018 nm$, and $ 2517 \times 2335$ pixels with a spatial resolution of $2.5 m$. As shown in Fig \ref{fig1}, there are 19 classes of land cover, including urban and rural areas. 

\begin{figure}[htbp]
	\centering
	\includegraphics[width=7.5cm]{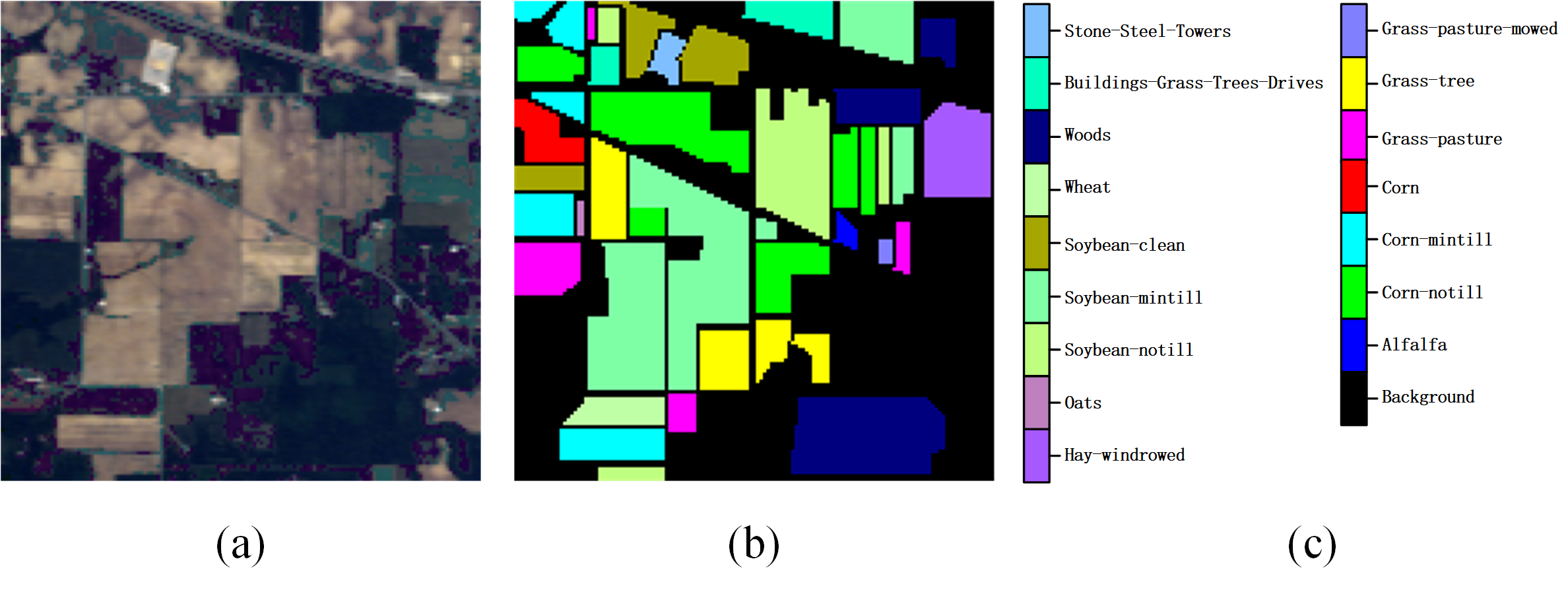}
	\caption{ The  color map, label map and label color of IP dataset}\label{fig2}
\end{figure}

\textbf{Indian Pines (IP)}: It was imaged by an airborne visible infrared imaging spectrometer (AVIRIS) in Indiana, USA. This data contains 200 bands with the wavelength ranging from $0.4-2.5(10^{-6})m $. The size is $145\times145$ pixels, with a spatial resolution of about $ 20m$. As shown in Fig \ref{fig2}, there are 16 classes of land cover  including crops and natural vegetation.

\begin{figure}[htbp]
	\centering
	\includegraphics[width=7cm]{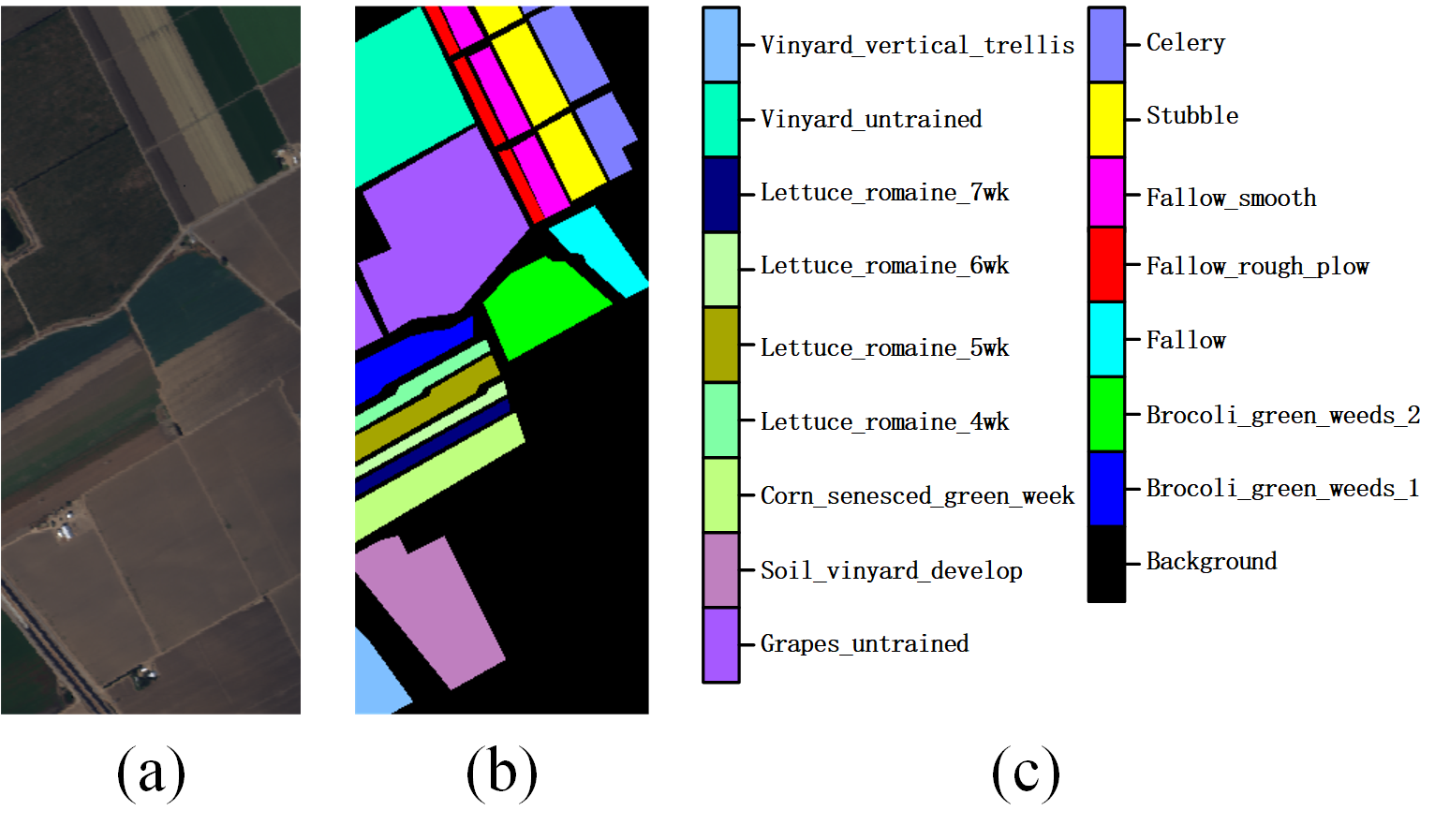}
	\caption{ The color map, label map and label color of SA dataset}\label{fig3}
\end{figure}

\textbf{Salinas (SA)}: This dataset was captured  by the airborne visible infrared imaging spectrometer (AVIRIS) in the Salinas Valley, California, USA. It contains 204 wavebands with a size of $512 \times 217$ pixels, and a spatial resolution of about $3.7m$. As shown in Fig \ref{fig3}, there are 16 classes of land cover, which include  vegetables, exposed soil, etc. 

\begin{figure}[htbp]
	\centering
	\includegraphics[width=5cm]{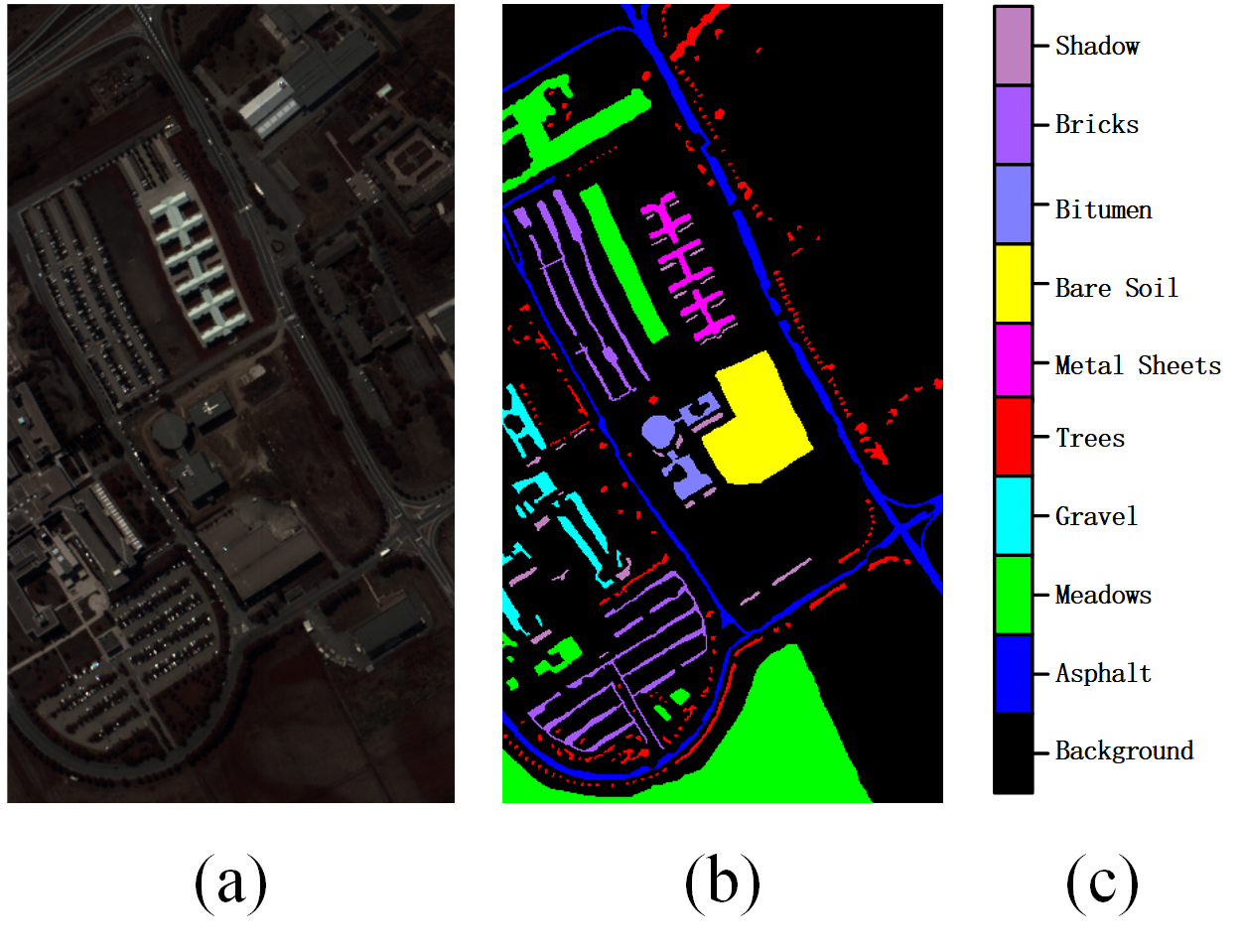}
	\caption{The  color map, label map and label color of PU dataset }\label{fig4}
\end{figure}

\textbf{Pavia University (PU)}: This dataset was imaged by the German airborne reflective optical spectral imager (ROSIS) in the city of Pavia, Italy. The data contains 103 wavebands, with a size of $610 \times 340$ pixels and a spatial resolution of about $1.3m$. As shown in Fig \ref{fig4}, there are 9 classes of land cover, including trees, asphalt roads, bricks, etc. 

\subsubsection{Metric}
Also for the fairness of comparison,  we adopt the metrics of overall accuracy rate (OA), average accuracy rate (AA) and kappa coefficient which have been widely used to evaluate the performance of proposed method. The definitions of these metrics can be referred to the work of  \citep{26}.

\subsubsection{Implementation and Configuration}

The patches with the size of $9\times9$ pixels centering around each pixel  are sampled from hyperspectral images and taken  as input of the model. All linear and convolutional layers are normalized using Xavier, and the Adam optimizer is used to optimize  the model. The learning rate is 0.001, and 10000 episodes are there. For the weights $\lambda_i$($i=1,2,3$)  shown in Eq.(7), they are set to 10, 1 and 1 for IP dataset, 10, 2 and 0.005 for PU dataset, and 10, 2 and 0.005 for SA dataset. The average and deviation of AA, OA and  kappa over 10 random seeds are reported. Moreover,  because there  are too few labeled samples  available for training in the target domain,  the paper also uses Gaussian noise to enhance the data. For each class, there are 200 samples enhanced from 5 samples.

\subsection{Comparison with Related Methods}
In order to verify the performance of the proposed method, we select several typical classification methods for comparison, including SVM \citep{3}, 3D-CNN\citep{57}, DAAN\citep{58}, DSAN\citep{59}, DFSL\citep{60}, DCFSL\citep{25}, CMFSL\citep{61} and Gia-CFSL\citep{48}.  Among them, SVM and 3D-CNN are general supervised learning models that require consistent dimensions and classes of training and testing data, so only labeled samples in the target domains are used for training and other unlabeled samples are for testing in our experiments. DAAN and DSAN are classification models based on transfer learning, which use all labeled  data of source domain and target domain for transfer training.  DFSL is a classic metric-based few-shot learning method which uses  the Euclidean distance metric for inferring the classes of samples.  Building on DFSL method, DCFSL  utilizes  a discriminator to align sample features from different domains, to address hyperspectral image classification under cross-domain setting. In addition, CMFSL learns global class representations for each training episode by interactively using training samples from the base and novel classes. Gia-CFSL addresses the cross-domain few-shot hyperspectral image classification through graph alignment.

\begin{table*}[h]
	\caption{Comparison results on IP dataset (5 labeled samples per class)}\label{IP-result}
	\setlength{\tabcolsep}{1.8mm}{
		\begin{tabular}{@{}llllllllll@{}}
			\toprule
			\textbf{Class} & \textbf{SVM}    & \textbf{3D-CNN}      & \textbf{DAAN}     & \textbf{DSAN}      & \textbf{DFSL}   & \textbf{DCFSL} & \textbf{CMFSL} &  \textbf{Gia-CFSL}  & \textbf{ours}   \\ \midrule
			1& 72.20&95.12	&78.05&	\textbf{100.00}&	96.75&	96.34 & 96.34 & 92.44 &	98.05 \\ 
			2&	34.27&	37.70&	21.36&	35.49&	36.38&	48.10 & 48.70 & 43.08 &	\textbf{51.78}  \\ 
			3&	39.18&	19.77&	10.79&	59.15&	38.34&	53.27 & \textbf{59.96} & 51.24 &	55.43 \\ 
			4&	50.34&	32.51&	65.09&	\textbf{83.19}&	77.16&	81.59 & 78.88 & 75.60 &	77.28 \\ 
			5&	69.75&	\textbf{88.45}&	62.97&	46.44&	73.92&	73.77 & 77.13 & 70.98 &	80.61 \\ 
			6&	66.36&	73.65&	71.72&	85.93&	86.25&	86.14 & 80.36 & 83.08 &	\textbf{91.26} \\ 
			7&	89.13&	81.82&	\textbf{100.00}&	\textbf{100.00}&	97.10&	99.57 & \textbf{100.00} & 99.57 & 98.7 \\ 
			8&	68.73&	53.35&	\textbf{97.46}&	67.23&	81.82&	82.98 & 90.53 & 85.48 &	83.07 \\ 
			9&	86.67&	\textbf{100.00}&	93.33&	\textbf{100.00}&	75.56&\textbf{100.00} & \textbf{100.00} & \textbf{100.00} &	98.67 \\ 
			10&	37.49&	41.35&	50.78&	63.91&	52.22&	63.45 & 63.52 & 64.38 &	\textbf{66.61} \\
			11&	33.96&	66.71&	\textbf{69.63}&	52.86&	59.96&	60.09 & 60.65 & 63.96 &	63.53 \\ 
			12&	31.43&	37.40&	37.59&	47.28&	36.56&	44.73 & 48.16 & 50.49 &	\textbf{53.91} \\
			13&	86.50&	85.71&	\textbf{100.00}&	97.00&	98.00&	99.00 & 99.35 & 97.95 &	99.05 \\
			14&	62.93&	62.57&	\textbf{95.56}&	70.71&	84.63&	81.72 & 81.33 & 81.94 &	83.76 \\
			15&	28.08&	56.42&	18.37&	\textbf{87.93}&	74.10&	71.44 & 72.31 & 64.51 &	82.10 \\
			16&	90.91&	90.36	&\textbf{100.00}&	\textbf{100.00}&	\textbf{100.00}	&	98.86 & 99.66 & 99.32 &	\textbf{100.00} \\ \midrule
			OA & 45.85 & 54.76  & 57.77 & 60.25&  61.69&  65.85 & 66.85 & 65.70 &   \textbf{69.46}\\ 
			AA & 59.24 &  63.93 & 67.04&   74.82  &73.05 & 77.57 & 78.56 & 76.50 &   \textbf{80.24}\\ 
			Kappa & 39.68 & 48.72&  52.16  & 55.75  &56.78  &  61.59   & 62.67 & 61.28 & \textbf{65.69} \\ \bottomrule
			
		\end{tabular}
	}
\end{table*}

\begin{figure*}[h]
	\centering
	\includegraphics[width=12cm]{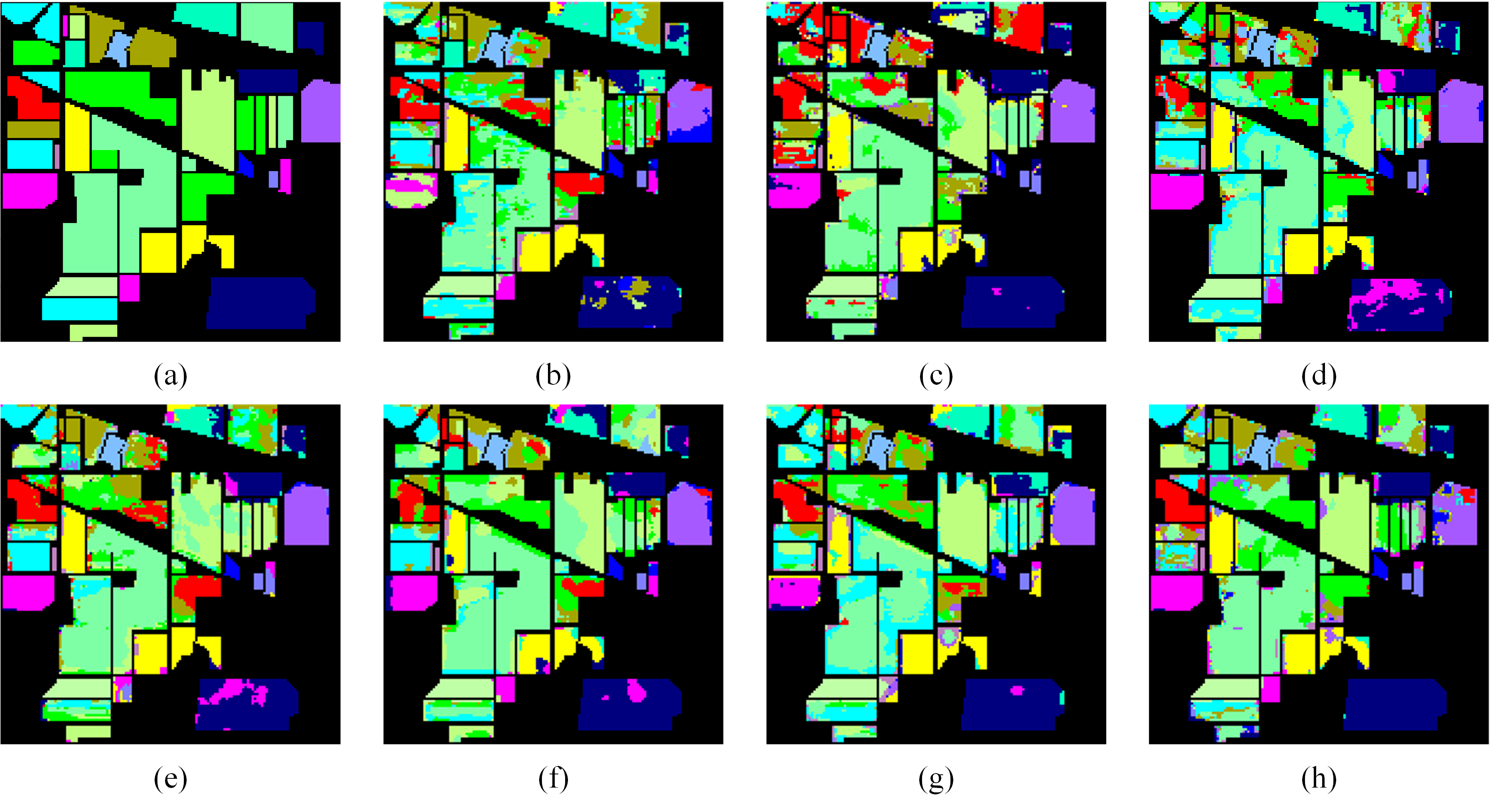}
	\caption{Data visualization and classification diagrams of target scenes obtained using different methods on IP dataset,  include: (a) ground reality map, (b) DAAN (57.77\%), (c) DSAN (60.25\%), (d) DFSL (61.69\%),  (e) DCFSL (65.85\%), (f) CMFSL (66.85\%), (g) Gia-CFSL (65.70\%), and (h) ours (69.46\%).}\label{fig5}
\end{figure*}

\begin{table*}[h]
	\caption{Comparison results on PU dataset (5 labeled samples per class)}\label{PU-result}
	\setlength{\tabcolsep}{2mm}{
		\begin{tabular}{@{}llllllllll@{}}
			\toprule
			\textbf{Class}& \textbf{SVM}    & \textbf{3D-CNN}      & \textbf{DAAN}     & \textbf{DSAN}      & \textbf{DFSL}   & \textbf{DCFSL} & \textbf{CMFSL} &  \textbf{Gia-CFSL} & \textbf{ours}   \\ \midrule
			1&	\textbf{88.98}&	59.82&	65.44&	70.48&	73.43&	79.07 & 81.50 & 78.23 &	78.77\\ 
			2&	83.91&	63.05&  86.56&  83.84&	\textbf{89.25}&	85.43 & 83.45 & 87.73 &	83.54\\
			3&	39.98&	\textbf{68.91}&	56.69&	35.96&	48.09&	62.82 & 68.72 & 63.13 &	68.83\\
			4&	60.22&	77.31&	77.93&	\textbf{95.36}&	84.72&	92.69 & 90.14 & 91.41 &	93.11\\
			5&	95.44&	90.77&	97.39&	95.37&	99.65&	99.47 & \textbf{99.71} & 99.24 &	\textbf{99.71}\\
			6&	37.12&	63.40&	47.33&	50.06&	67.81&	74.82 & 76.04 & 72.36 &	\textbf{79.72}\\ 
			7&	40.62&	87.64&	\textbf{90.87}&	84.83&	64.48&	75.12 & 80.96 & 76.83 &	89.26\\ 
			8&	68.17&	57.27&	55.51&	58.01&	67.37&	65.22 & \textbf{77.57} & 70.85 &	73.39\\ 
			9&	99.13&	95.57&	\textbf{99.26}&	97.45&	92.92&	98.45 & 98.13 & 98.27 &	96.36\\  \midrule
			OA & 64.12 & 65.74  & 74.67 & 74.74&  79.63&  81.28 & 82.01 & 82.31 &   \textbf{82.41} \\ 
			AA & 68.18 &  73.72 & 75.22&  74.60  &76.41& 81.46 & 82.29 & 84.02 &   \textbf{84.74} \\ 
			Kappa & 55.59 & 57.37&  67.03  & 67.13 &73.05  & 75.74   & 77.05 & 76.92 &\textbf{77.40} \\  \bottomrule
			
		\end{tabular}
	}
\end{table*}

\begin{figure*}
	\centering
	\includegraphics[width=12cm]{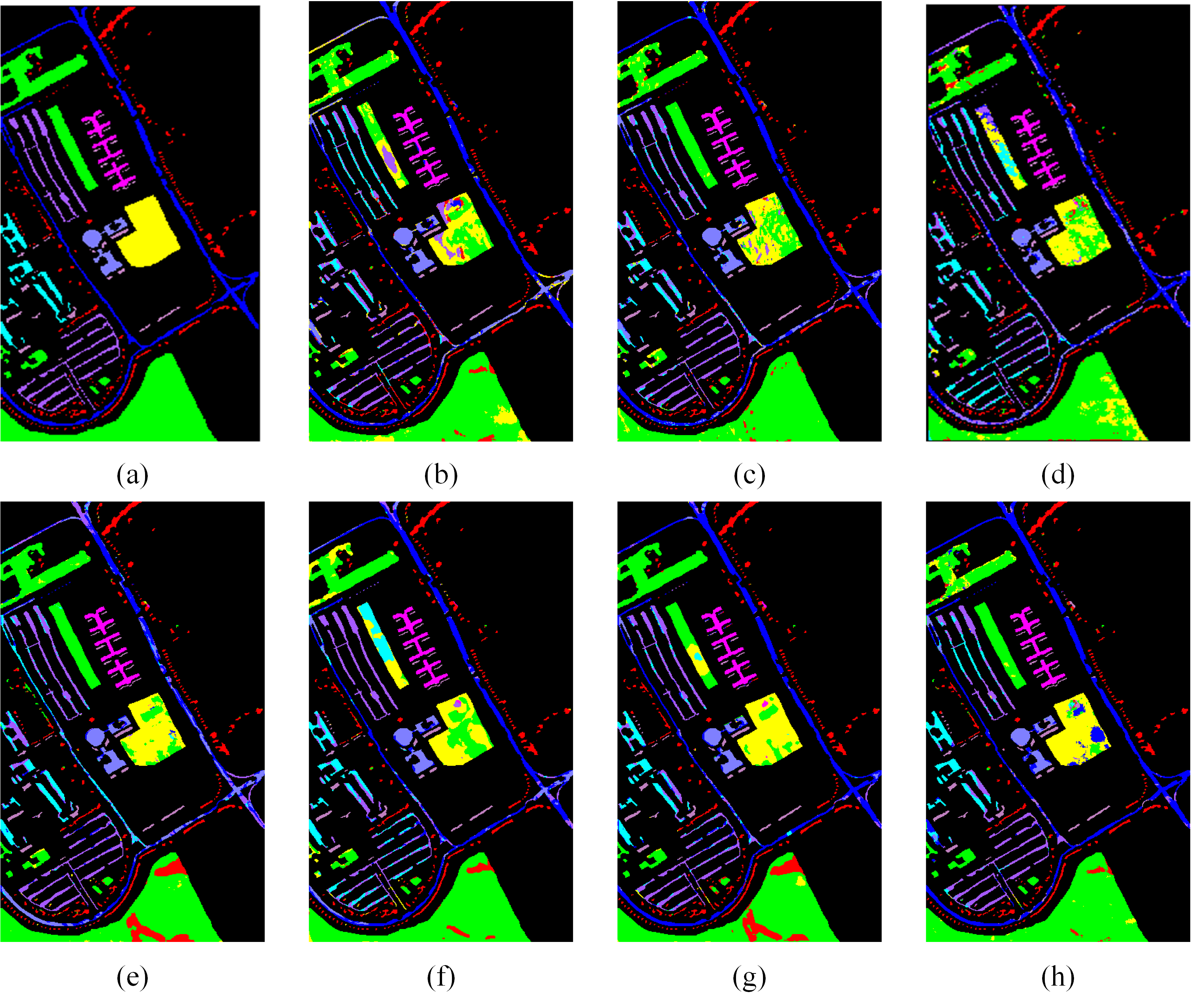}
	\caption{Data visualization and classification diagrams of target scenes obtained using different methods on PU dataset, including: (a) ground reality map, (b) DAAN (74.67\%), (c) DSAN (74.74\%), (d) DFSL (79.63\%), (e) DCFSL (81.28\%), (f) CMFSL (82.01\%), (g) Gia-CFSL (82.31\%), and (h) ours (82.41\%).}\label{fig6}
\end{figure*}

\begin{table*}
	\caption{Comparison results on SA dataset (5 labeled samples per class)}\label{SA-result}
	\setlength{\tabcolsep}{2mm}{
		\begin{tabular}{@{}llllllllll@{}}
			\toprule
			\textbf{Class} & \textbf{SVM}    & \textbf{3D-CNN}      & \textbf{DAAN}     & \textbf{DSAN}      & \textbf{DFSL}   & \textbf{DCFSL} & \textbf{CMFSL} &  \textbf{Gia-CFSL} & \textbf{ours}   \\ \midrule
			1 &	97.57 &	95.29 &	98.30 &	99.20 &	73.92  & 99.27  & 97.13 & 99.06 &	\textbf{99.65} \\ 
			2 &	87.43 &	97.20 &	97.02 &	99.79 &	96.85 &	98.89  & 99.16 & 99.34 &	\textbf{99.97} \\ 
			3 &	82.95 &	91.45 &	95.94 &	83.11 &	\textbf{96.28} &	93.21  & 90.72 & 89.88 &	89.37 \\ 
			4 &	99.11 &	97.31 &	\textbf{99.93} &	\textbf{99.93} &	99.11 &	99.52  & 99.17 & 98.76 &	99.47 \\ 
			5 &	94.29 &	91.24 &	\textbf{97.04} &	89.97 &	80.72 &	91.26  & 93.03 & 89.16 &	91.18 \\ 
			6 &	98.36 &	98.80 &	\textbf{100.00} &\textbf{100.00} & 91.63 & 99.38  & 99.46 & 97.63 & 99.93 \\ 
			7 &	94.39 &	99.69 &	\textbf{99.86} &	98.49 &	97.73 &	99.40  & 97.73 & 99.29 &	99.06 \\ 
			8 &	59.99 &	66.40 &	57.35 &	67.46 &	\textbf{82.33} &	75.83  & 70.05 & 76.80 &	79.85 \\ 
			9 &	96.09 &	96.25 &	99.82 &	98.42 &	94.44 &	98.70  & 99.20 & 98.32 &	\textbf{99.90} \\
			10&	71.45 &	70.72 &	\textbf{93.55} &	89.06 &	80.96 &	84.91  & 84.27 & 81.31 &	87.17 \\
			11&	91.25 &	93.15 &	89.18 &	\textbf{98.68} &	93.38 &	98.14  & 96.69 & 96.43 &	98.67 \\
			12&	97.22 &	99.65 &	98.60 &	99.27 &	97.94 &	\textbf{99.78}  & 98.36 & 98.76 &	99.27 \\
			13&	97.30 &	92.63 &	\textbf{99.67} &	99.23 &	95.79 &	99.15  & \textbf{99.67} & 98.11 &	98.4 \\
			14&	91.84 &	93.56 &	95.49 &	\textbf{99.34} &	98.87 &	98.59  & 98.81 & 97.89 &	98.78 \\
			15&	60.52 &	68.02 &	52.42 &	48.80 &	71.13 &74.36  & 77.11 & 74.89 & \textbf{78.08} \\
			16& 81.45 &	81.41 & 97.72 &	\textbf{99.11} &	90.57 &	89.51  & 87.08 & 82.14 &	92.5 \\ \midrule
			OA & 80.71 & 84.20  & 83.30     & 84.13   &  86.95 & 89.14 & 87.99 & 88.45 &   \textbf{90.79}\\ 
			AA & 87.58 & 89.56 & 91.99      &   91.87     & 90.08  &  93.74   & 92.98 & 92.36 &\textbf{94.77}\\ 
			Kappa & 78.61 & 82.46&  81.44  & 82.36    &85.51 &  87.94   & 86.67 & 87.16 &\textbf{89.69} \\ \bottomrule
		\end{tabular}
	}
\end{table*}

\begin{figure*}
	\centering
	\includegraphics[width=12cm]{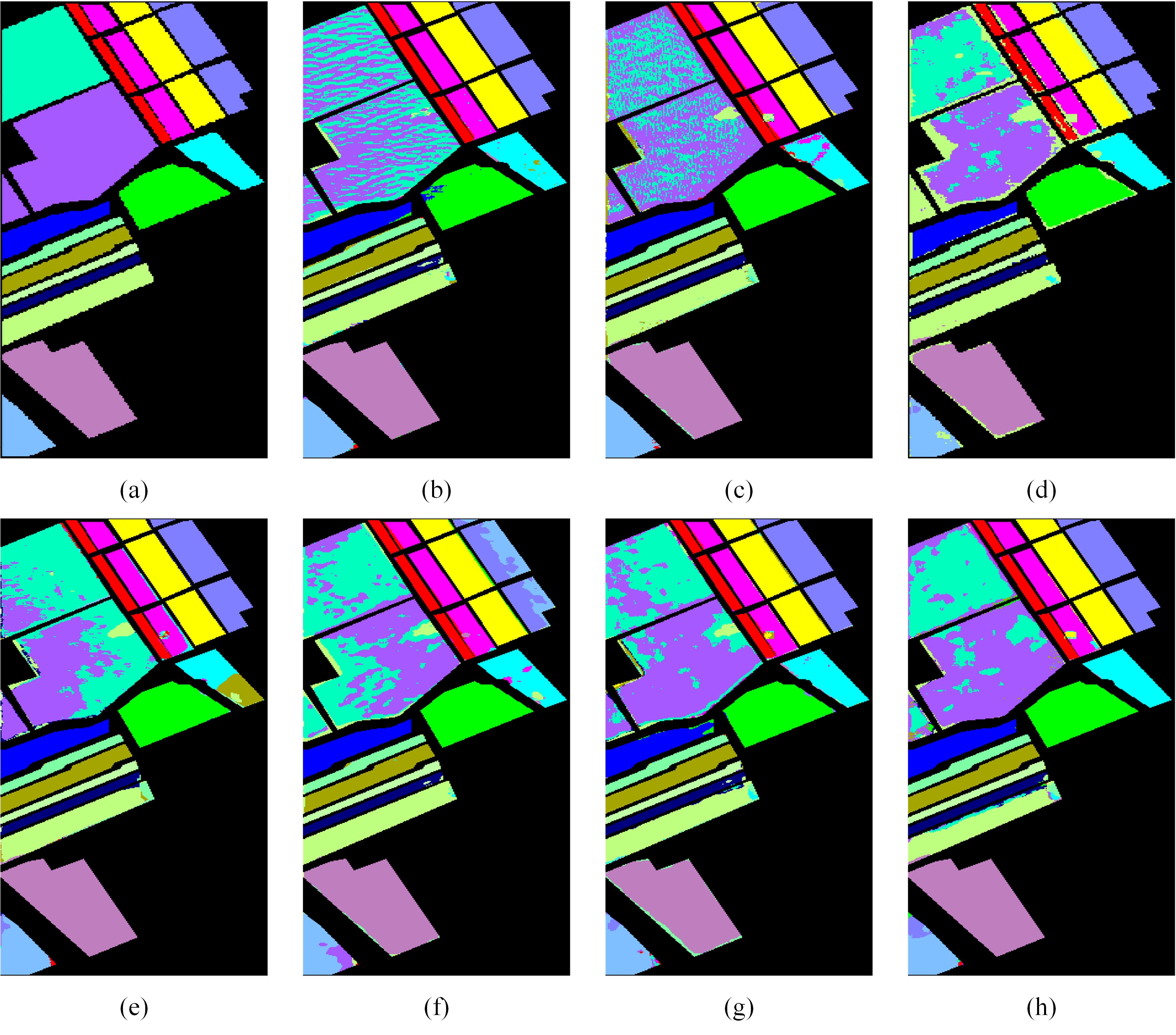}
	\caption{Data visualization and classification diagrams of target scenarios obtained using different methods on SA dataset, including: (a) Ground Realistic Map, (b) DAAN (83.30\%), (c) DSAN (84.13\%), (d) DFSL (86.95\%), (e) DCFSL (89.14\%), (f) CMFSL (87.99\%), (g) Gia-CFSL (88.45\%), and (h) ours (90.79\%).}\label{fig7}
\end{figure*}

Table \ref{IP-result} to \ref{SA-result} reports the specific  classification results over each class and the AA, OA, and kappa values of each model on the datasets of three target domains.  From the comparison results, it can be seen that compared with SVM and 3DCNN which only use a small number of labeled samples from target domain for training,  there are significant improvements for all the other methods which have  used the samples from source domain to participate in training. This shows that when there are few labeled samples in the target domain, using the samples from source domain for training can help improve the classification accuracy on  target domain.

Moreover,  among these methods of DAAN, DSAN, DFSL, DCFSL, CMFSL, Gia-CFSL and ours which use the data from both source domain and  target domain for training, the results of these few-shot learning methods are better than those of non few-shot learning methods. In Table \ref{IP-result} to \ref{SA-result}, the OA of the few-shot learning methods including DFSL, DCFSL, CMFSL, Gia-CFSL and ours exceeds the non few-shot learning methods of DAAN and DSAN by $2\%$ to $8\%$ on IP, PU, and SA datasets. This indicates that few-shot learning methods have more advantages  when addressing  classification problems with few labeled samples. The episode-based training method adopted by these  few-shot learning methods enables that the model trained  has stronger generalization ability and makes better use of  knowledge from source domain. Further, compared with the selected few-shot learning methods, the proposed method in this paper has shown superior performance.  For example,   the OA of our method outperforms other few-shot learning methods on IP dataset by  $3\%$. These results indicate the effectiveness of the proposed relation learning modules for hyperspectral image classification. 

The classification maps corresponding to each method are shown in Fig \ref{fig5} to \ref{fig7}. Since the SVM and 3D-CNN results are significantly weaker than other methods, their classification results are not presented.  In these maps,  color pixels are the ones which have been marked,  and the black pixels are the ones which are unlabeled and shown as  backgrounds. By observing the results shown in the classification diagrams,  it can be found that the classification map generated by the proposed method in this paper is most similar to the real ground map. These results also confirm that the method in this paper effectively improves the classification accuracy of hyperspectral image classification under cross-domain few-shot setting.

\subsection{Feature Visualization}

\begin{figure*}
	\centering
	\includegraphics[width=16cm]{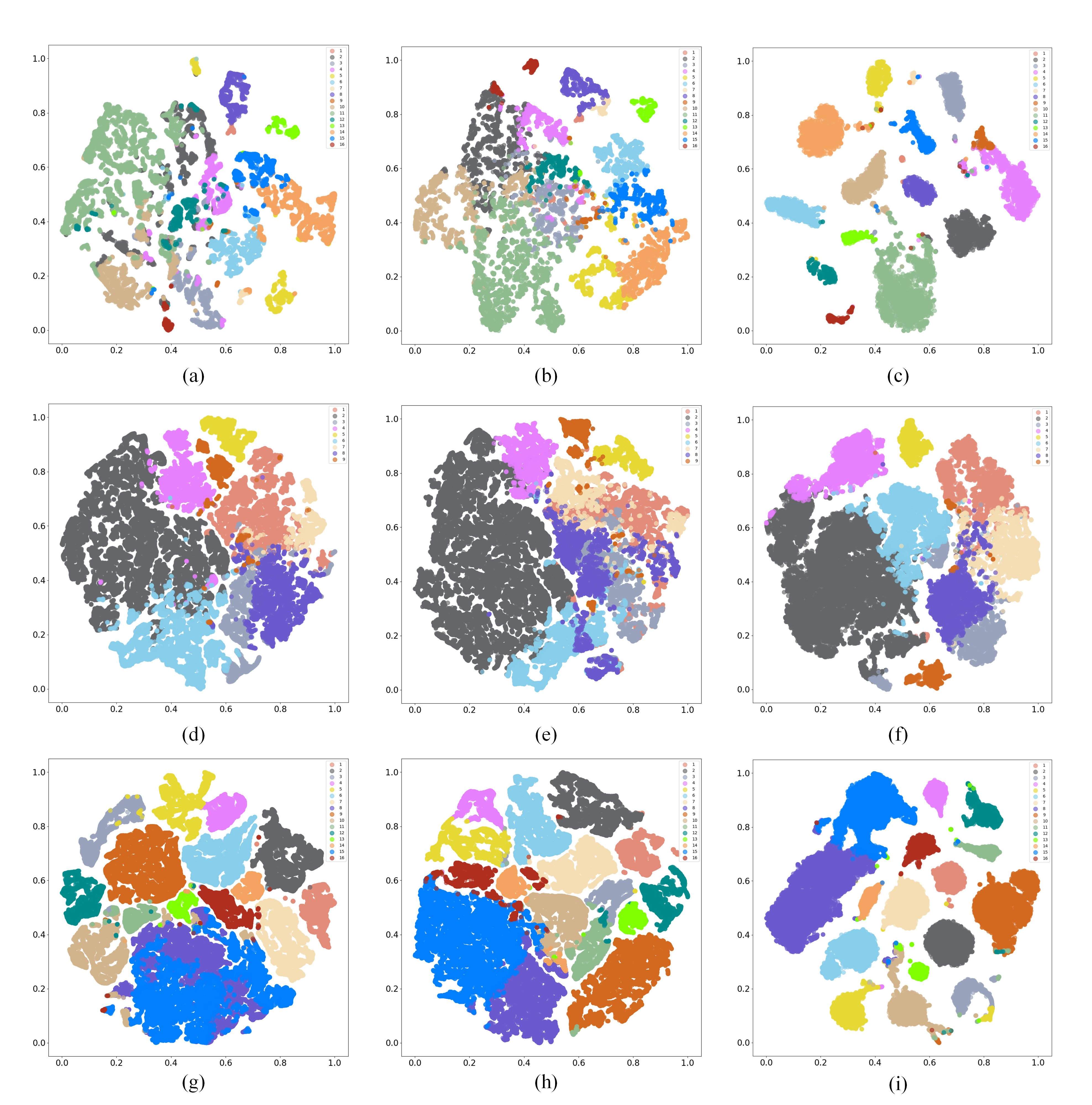}
	\caption{2-D feature visualization on three datasets. (a) CMFSL features on IP. (b) Gia-CFSL features on IP. (c) ours features on IP. (d) CMFSL features on PU. (e) Gia-CFSL features on PU. (f) ours features on PU. (g) CMFSL features on SA. (h)Gia-CFSL features on SA. (i) ours features on SA.}\label{sne}
\end{figure*}

In order to intuitively reflect the feature extraction performance of different methods, we use the t-distributed stochastic neighbor embedding (t-SNE) technique to show the 2-D projected features of CMFSL\citep{61}, Gia-CFSL\citep{48}, and ours on three datasets in Figure \ref{sne}.

The figure shows the classification performance of three methods on IP, PU, and SA datasets, respectively. We can observe that by our proposed method, the distance between samples of the same class is shortened, while the distance between samples of different classes is increased. This indicates the contributes of the proposed relation learning modules.

\subsection{Ablation Experiments}
This paper proposes to learn the relations between samples and take them into model learning process to improve the hyperspectral image classification performance. Besides the domain-level domain discrimination, we have proposed the class-level contrastive learning module and the set-level cross-attention learning module.  In order to evaluate the contribution of these modules,  the ablation experiment has been done, and the results are shown in Table \ref{ablation-result}. Different variants have been constructed by excluding some of these modules to validate the benefits brought by them. These variants can be seen from the different rows of Table \ref{ablation-result}. 

The variant $v_0$ is the model we proposed. The variant of $v_1$ represents the model without the class-level contrastive learning module module, and the  variant of $v_2$ is that without the  set-level cross-attention learning module. From Table \ref{ablation-result}, it can be seen that the performance of the complete model, i.e.,  $v_0$,  is superior to other types of ablation models in terms of OA. For example, the performance of the proposed model reached $69.46\%$ in OA. But it falls down to $66.43\%$ when removing the class-level contrastive learning module, and to $67.49\%$ when removing the set-level cross-attention learning module. This ablation result further confirms the effectiveness of the proposed relation learning modules.

\begin{table*}
	\caption{The results of ablation experiment in terms of OA}\label{ablation-result}
	\setlength{\tabcolsep}{3mm}{
		\begin{tabular}{@{}lll|lll@{}}
			\toprule
			Variants & Cross-attention & Contrastive learning & IP & PU  & SA\\
			\midrule
			$v_0$ &\checkmark & \checkmark &69.46 +- 3.09   &82.41 +- 3.70 &90.79 +- 1.24   \\
			$v_1$ &\checkmark & $\times$  &66.43 +- 2.32  &82.08 +- 3.96    &88.73 +- 2.53 \\
			$v_2$ & $\times$  &  \checkmark  &67.49 +- 2.24 &82.13 +- 4.96 &90.44 +- 1.97\\
			\bottomrule
		\end{tabular}
	}
\end{table*}

\subsection{Sensitive Analysis of Parameters}
There are several parameters which affect the performance of the proposed method. We have done the experiments to study how sensitive the proposed method is to these parameters. In this section, we report the analysis results of three kinds of parameters, namely, the number of shots, the temperature $\tau$ used in the contrastive loss, and the dimension of mapping module. 

\subsubsection{Effect of the Number of Shots}
In order to analyze the effect of the number of labeled training samples from target domains on classification performance, we have set the number of labeled samples  in the set of $[5, 10, 15, 20, 25]$ for testing. The performance changes  of the proposed method on three datasets are shown in Table \ref{shot-result}. It can be seen that as the number of labeled samples increases, the performance in terms of OA, AA, and kappa  improves significantly. This is because more labeled samples for training mean that more knowledge is there which are helpful for predicting the classes of query samples. Because  the number of  pixels of  Oats class in  IP dataset is 20,  the experiments that set the number of shots to more than 20 have not been carried out. 

\begin{table*}[h]
	\caption{The classification results of proposed method under different number of labeled samples per class from target domain (N is the number of labeled samples per class)\label{tab:table1}}\label{shot-result}
	\setlength{\tabcolsep}{4.5mm}{
		\begin{tabular}{@{}llllll@{}}
			\toprule
			N & 5 & 10 & 15 & 20 & 25  \\
			\midrule
			Dataset  &	\multicolumn{5}{c}{IP} \\
			\midrule
			OA & 69.46 +- 3.09 & 80.36 +- 2.95 & 84.39 +- 3.03 & - &-\\
			AA & 80.24 +- 2.09 & 88.65 +- 1.29& 91.32 +- 1.43 & - &-\\
			Kappa &65.69 +- 3.30 & 77.86 +- 3.25 & 82.36 +- 3.28 & - &-\\
			\midrule
			Dataset  &	\multicolumn{5}{c}{PU} \\
			\midrule
			OA & 82.41 +- 3.70& 87.63 +- 3.00 & 90.01 +- 3.61 & 93.12 +- 2.05 &94.15 +- 1.51\\
			AA & 84.74 +- 1.89 &89.37 +- 1.37 & 92.28 +- 1.19&94.22 +- 1.25 &94.97 +- 0.86 \\
			Kappa & 77.40 +- 4.32 & 83.98 +- 3.67 &87.09 +- 4.37&91.00 +- 2.60 &92.34 +- 1.90\\
			\midrule
			Dataset  &	\multicolumn{5}{c}{SA} \\
			\midrule
			OA & 90.79 +- 1.24 &92.61 +- 1.65&94.14 +- 0.68& 94.83 +- 0.38 &95.07 +- 0.61\\
			AA &94.45 +- 0.90&96.31 +- 0.57&97.14 +- 0.40 &97.59 +- 0.25 &97.80 +- 0.43\\
			Kappa & 89.76 +- 1.38 & 91.79 +- 1.83& 93.48 +- 0.76&94.25 +- 0.42 &94.51 +- 0.68\\
			\bottomrule
		\end{tabular}
	}
\end{table*}

\subsubsection{Effect of the Temperature $\tau$ Used in  Contrastive Loss}

\begin{figure*}[h]
	\centering
	\includegraphics[width=\textwidth]{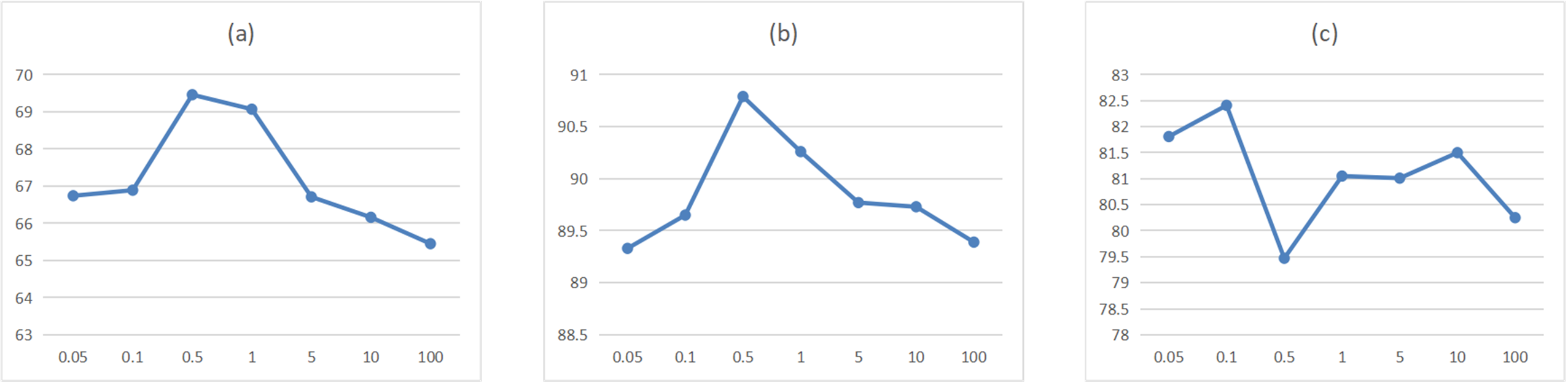}
	\caption{The performance changes in terms of OA on IP (a), SA (b), PU (c) datasets when changing $\tau$  (5 labeled samples)}\label{fig11}
\end{figure*}

To illustrate the effects of $\tau$ on the classification performance, we tested the values in $[0.05,  0.1, 0.5, 1, 10, 100]$. The performance changes of the proposed method on three datasets are shown in Figure \ref{fig11}.  The figure shows that as the change of  $\tau$, the optimal performance is achieved when $\tau=0.5$ on  IP and SA datasets, and  $\tau=0.1$ for PU dataset. These differences may be attributed to that the IP, SA, and PU datasets are imaged by different spectrometers.  According to these results, we have set $\tau$ to 0.5 for  IP and SA datasets and 0.1 for PU dataset in our experiments.

\subsubsection{Effect of the Dimension of  Mapping Module}
The hyperspectral images from different domains may contain different number of bands. So, the mapping modules are adopted for transforming the input hyperspectral images into the new ones with desired number of bands.  We believe that when there are more bands for the transformed images, more information or features of the original images are kept, which contributes to the following calssification tasks. To test the number of bands of transformed images, i.e., the  dimension of the mapping module, we set the  dimension to be 50, 75, 100, 125 and 150, respectively. The final performance of the proposed model on three datasets is shown in Table \ref{mapping-result}.  The results show that as the dimension of the mapping module increases, the performance in terms of OA, AA, and kappa are improved to some extent. It proves that  higher dimension of the mapping module contributes to the classification tasks. But higher dimension means that there are more computation. So, it is better to adopt a compromise for the dimension of the mapping module.

\begin{table*}[h]
	\caption{The classification results of the proposed method under different  dimensions of the mapping modules}\label{mapping-result}
	\setlength{\tabcolsep}{4mm}{
		\begin{tabular}[\textwidth]{@{}llllll@{}}
			\toprule
			Dimension	& 50 & 75 & 100 & 125 & 150  \\
			\midrule
			Dataset  &	\multicolumn{5}{c}{IP} \\
			\midrule
			OA & 68.71 +- 2.37 & 70.01 +- 2.18 & 69.46 +- 3.09 & 70.38 +- 3.48 &70.35 +- 3.04\\
			AA & 80.07 +- 1.51 & 81.27 +- 0.83& 80.24 +- 2.09 & 81.56 +- 1.49 &81.39 +- 1.72\\
			Kappa & 64.79 +- 2.41 & 66.29 +- 2.22 & 65.69 +- 3.30 & 66.79 +- 3.60 &66.56 +- 3.39\\
			\midrule
			Dataset  &	\multicolumn{5}{c}{PU} \\
			\midrule
			OA & 80.51 +- 4.50 & 81.59 +- 5.16 & 82.41 +- 3.70 & 82.13 +- 5.39 &82.99 +- 4.51\\
			AA & 81.54 +- 1.75 & 82.93 +- 2.81 & 84.74 +- 1.89 & 84.58 +- 1.97 &85.17 +- 1.98 \\
			Kappa & 74.95 +- 5.16 & 76.29 +- 5.98 & 77.40 +- 4.32 &77.14 +- 6.08 &78.16 +- 5.18\\
			\midrule
			Dataset  &	\multicolumn{5}{c}{SA} \\
			\midrule
			OA & 90.32 +- 1.80 &90.35 +- 2.26 & 90.79 +- 1.24 & 90.71 +- 1.64 &90.18 +- 1.44\\
			AA & 94.66 +- 1.04 &94.58 +- 1.11 & 94.45 +- 0.90 &94.65 +- 1.28 &94.46 +- 1.20\\
			Kappa & 89.25 +- 1.98 & 89.28 +- 2.49 & 89.76 +- 1.38 &89.67 +- 1.83 &89.09 +- 1.60\\
			\bottomrule
		\end{tabular}
	}
\end{table*}

\section{Conclusion}

Taking the relations between samples into consideration, this paper proposes to learn the sample relations from multi-level and use them to constrain the model learning process, to improve the few-shot hyperspectral image classification performance. Focusing on the cross-domain few-shot hyperspectral image classification, three kinds of similarity relations have been addressed from the levels of cross-domain, cross-set and cross-class. Accordingly, three modules including the domain-level domain discriminator, the set-level cross-attention learning module and the class-level contrastive learning module are presented to follow the feature extractor to learn the relations between samples in embedding space.  Extensive experiments have been done, and the results have shown the  contribution of these relation learning modules for the cross-domain few-shot hyperspectral image classification. 

Frankly speaking, this paper has  included only three kinds of sample relations. There may be more kinds of sample relations which are also valuable for investigation, for example, the contrastive relations between different domains, the relations between different pixeles within one patch, and the relations among the pixels of query and support patches. Moreover, taking the labeled samples from source domain into training has shown positive effects. However, it may be also expensive to obtain large number of labeled samples from the source domain for hyperspectral image classification. In this case, it is also a good direction for exploration to adopt the unsupervised learning model such as variational autoencoder to pre-train the hyperspectral image feature extractor. 


\section{Declarations}

\begin{itemize}
	\item Conflict of interest The authors declare no conflict of interest.
	\item Ethics approval Not applicable.
	\item Availability of data and materials The datasets generated and analysed during the current study are available at \url{https://www.ehu.eus/ccwintco/index.php/Hyperspectral_Remote_Sensing_Scenes} and \url{http://naotoyokoya.com/Download.html}.
	\item Code availability The codes are available at \url{https://github.com/HENULWY/STBDIP}.
	\item Authors' contributions Chun Liu drafted the manuscript. Longwei Yang designed expriments. All authors have read and agreed to the published version of the manuscript.
\end{itemize}

\bibliographystyle{sn-mathphys.bst}
%
%

\begin{thebibliography}{58}
	\ifx \bisbn   \undefined \def \bisbn  #1{ISBN #1}\fi
	\ifx \binits  \undefined \def \binits#1{#1}\fi
	\ifx \bauthor  \undefined \def \bauthor#1{#1}\fi
	\ifx \batitle  \undefined \def \batitle#1{#1}\fi
	\ifx \bjtitle  \undefined \def \bjtitle#1{#1}\fi
	\ifx \bvolume  \undefined \def \bvolume#1{\textbf{#1}}\fi
	\ifx \byear  \undefined \def \byear#1{#1}\fi
	\ifx \bissue  \undefined \def \bissue#1{#1}\fi
	\ifx \bfpage  \undefined \def \bfpage#1{#1}\fi
	\ifx \blpage  \undefined \def \blpage #1{#1}\fi
	\ifx \burl  \undefined \def \burl#1{\textsf{#1}}\fi
	\ifx \doiurl  \undefined \def \doiurl#1{\url{https://doi.org/#1}}\fi
	\ifx \betal  \undefined \def \betal{\textit{et al.}}\fi
	\ifx \binstitute  \undefined \def \binstitute#1{#1}\fi
	\ifx \binstitutionaled  \undefined \def \binstitutionaled#1{#1}\fi
	\ifx \bctitle  \undefined \def \bctitle#1{#1}\fi
	\ifx \beditor  \undefined \def \beditor#1{#1}\fi
	\ifx \bpublisher  \undefined \def \bpublisher#1{#1}\fi
	\ifx \bbtitle  \undefined \def \bbtitle#1{#1}\fi
	\ifx \bedition  \undefined \def \bedition#1{#1}\fi
	\ifx \bseriesno  \undefined \def \bseriesno#1{#1}\fi
	\ifx \blocation  \undefined \def \blocation#1{#1}\fi
	\ifx \bsertitle  \undefined \def \bsertitle#1{#1}\fi
	\ifx \bsnm \undefined \def \bsnm#1{#1}\fi
	\ifx \bsuffix \undefined \def \bsuffix#1{#1}\fi
	\ifx \bparticle \undefined \def \bparticle#1{#1}\fi
	\ifx \barticle \undefined \def \barticle#1{#1}\fi
	\bibcommenthead
	\ifx \bconfdate \undefined \def \bconfdate #1{#1}\fi
	\ifx \botherref \undefined \def \botherref #1{#1}\fi
	\ifx \url \undefined \def \url#1{\textsf{#1}}\fi
	\ifx \bchapter \undefined \def \bchapter#1{#1}\fi
	\ifx \bbook \undefined \def \bbook#1{#1}\fi
	\ifx \bcomment \undefined \def \bcomment#1{#1}\fi
	\ifx \oauthor \undefined \def \oauthor#1{#1}\fi
	\ifx \citeauthoryear \undefined \def \citeauthoryear#1{#1}\fi
	\ifx \endbibitem  \undefined \def \endbibitem {}\fi
	\ifx \bconflocation  \undefined \def \bconflocation#1{#1}\fi
	\ifx \arxivurl  \undefined \def \arxivurl#1{\textsf{#1}}\fi
	\csname PreBibitemsHook\endcsname
	
	\bibitem[\protect\citeauthoryear{Li et~al.}{2019}]{1}
	\begin{barticle}
		\bauthor{\bsnm{Li}, \binits{S.}},
		\bauthor{\bsnm{Song}, \binits{W.}},
		\bauthor{\bsnm{Fang}, \binits{L.}},
		\bauthor{\bsnm{Chen}, \binits{Y.}},
		\bauthor{\bsnm{Benediktsson}, \binits{J.A.}}:
		\batitle{Deep learning for hyperspectral image classification: An overview}.
		\bjtitle{IEEE Transactions on Geoscience and Remote Sensing}
		\bvolume{PP}(\bissue{99}),
		\bfpage{1}--\blpage{20}
		(\byear{2019})
	\end{barticle}
	\endbibitem
	
	\bibitem[\protect\citeauthoryear{Jia et~al.}{2021}]{2}
	\begin{barticle}
		\bauthor{\bsnm{Jia}, \binits{S.}},
		\bauthor{\bsnm{Jiang}, \binits{S.}},
		\bauthor{\bsnm{Lin}, \binits{Z.}},
		\bauthor{\bsnm{Li}, \binits{N.}},
		\bauthor{\bsnm{Xu}, \binits{M.}},
		\bauthor{\bsnm{Yu}, \binits{S.}}:
		\batitle{A survey: Deep learning for hyperspectral image classification with
			few labeled samples}.
		\bjtitle{Neurocomputing}
		\bvolume{448},
		\bfpage{179}--\blpage{204}
		(\byear{2021})
	\end{barticle}
	\endbibitem
	
	\bibitem[\protect\citeauthoryear{Melgani and Bruzzone}{2004}]{3}
	\begin{barticle}
		\bauthor{\bsnm{Melgani}, \binits{F.}},
		\bauthor{\bsnm{Bruzzone}, \binits{L.}}:
		\batitle{Classification of hyperspectral remote sensing images with support
			vector machines}.
		\bjtitle{IEEE Transactions on Geoscience and Remote Sensing}
		\bvolume{42}(\bissue{8}),
		\bfpage{1778}--\blpage{1790}
		(\byear{2004})
	\end{barticle}
	\endbibitem
	
	\bibitem[\protect\citeauthoryear{Li et~al.}{2012}]{4}
	\begin{barticle}
		\bauthor{\bsnm{Li}, \binits{J.}},
		\bauthor{\bsnm{Bioucas-Dias}, \binits{J.M.}},
		\bauthor{\bsnm{Plaza}, \binits{A.}}:
		\batitle{Semisupervised hyperspectral image classification using soft sparse
			multinomial logistic regression}.
		\bjtitle{IEEE Geoscience and Remote Sensing Letters}
		\bvolume{10}(\bissue{2}),
		\bfpage{318}--\blpage{322}
		(\byear{2012})
	\end{barticle}
	\endbibitem
	
	\bibitem[\protect\citeauthoryear{Licciardi et~al.}{2011}]{5}
	\begin{barticle}
		\bauthor{\bsnm{Licciardi}, \binits{G.}},
		\bauthor{\bsnm{Marpu}, \binits{P.R.}},
		\bauthor{\bsnm{Chanussot}, \binits{J.}},
		\bauthor{\bsnm{Benediktsson}, \binits{J.A.}}:
		\batitle{Linear versus nonlinear pca for the classification of hyperspectral
			data based on the extended morphological profiles}.
		\bjtitle{IEEE Geoscience and Remote Sensing Letters}
		\bvolume{9}(\bissue{3}),
		\bfpage{447}--\blpage{451}
		(\byear{2011})
	\end{barticle}
	\endbibitem
	

	\bibitem[\protect\citeauthoryear{Zhang and Zheng}{2014}]{6}
	\begin{barticle}
		\bauthor{\bsnm{Zhang}, \binits{C.}},
		\bauthor{\bsnm{Zheng}, \binits{Y.}}:
		\batitle{Hyperspectral remote sensing image classification based on combined
			svm and lda}.
		\bjtitle{Multispectral, Hyperspectral, and Ultraspectral Remote Sensing
			Technology, Techniques and Applications V}
		\bvolume{9263},
		\bfpage{462}--\blpage{468}
		(\byear{2014})
	\end{barticle}
	\endbibitem


\color{black}{
	\bibitem[\protect\citeauthoryear{Qian et~al.}{2012}]{7}
	\begin{barticle}
		\bauthor{\bsnm{Qian}, \binits{Y.}},
		\bauthor{\bsnm{Ye}, \binits{M.}},
		\bauthor{\bsnm{Zhou}, \binits{J.}}:
		\batitle{Hyperspectral image classification based on structured sparse logistic
			regression and three-dimensional wavelet texture features}.
		\bjtitle{IEEE Transactions on Geoscience and Remote Sensing}
		\bvolume{51}(\bissue{4}),
		\bfpage{2276}--\blpage{2291}
		(\byear{2012})
	\end{barticle}
	\endbibitem
	
	\bibitem[\protect\citeauthoryear{Falco et~al.}{2015}]{8}
	\begin{barticle}
		\bauthor{\bsnm{Falco}, \binits{N.}},
		\bauthor{\bsnm{Benediktsson}, \binits{J.A.}},
		\bauthor{\bsnm{Bruzzone}, \binits{L.}}:
		\batitle{Spectral and spatial classification of hyperspectral images based on
			ica and reduced morphological attribute profiles}.
		\bjtitle{IEEE Transactions on Geoscience and Remote Sensing}
		\bvolume{53}(\bissue{11}),
		\bfpage{6223}--\blpage{6240}
		(\byear{2015})
	\end{barticle}
	\endbibitem
	
	\bibitem[\protect\citeauthoryear{Yu et~al.}{2017}]{9}
	\begin{barticle}
		\bauthor{\bsnm{Yu}, \binits{S.}},
		\bauthor{\bsnm{Jia}, \binits{S.}},
		\bauthor{\bsnm{Xu}, \binits{C.}}:
		\batitle{Convolutional neural networks for hyperspectral image classification}.
		\bjtitle{Neurocomputing}
		\bvolume{219},
		\bfpage{88}--\blpage{98}
		(\byear{2017})
	\end{barticle}
	\endbibitem
	
	\bibitem[\protect\citeauthoryear{Paoletti et~al.}{2019}]{10}
	\begin{barticle}
		\bauthor{\bsnm{Paoletti}, \binits{M.E.}},
		\bauthor{\bsnm{Haut}, \binits{J.M.}},
		\bauthor{\bsnm{Plaza}, \binits{J.}},
		\bauthor{\bsnm{Plaza}, \binits{A.}}:
		\batitle{Deep learning classifiers for hyperspectral imaging: A review}.
		\bjtitle{ISPRS Journal of Photogrammetry and Remote Sensing}
		\bvolume{158}(\bissue{Dec.}),
		\bfpage{279}--\blpage{317}
		(\byear{2019})
	\end{barticle}
	\endbibitem
	
	\bibitem[\protect\citeauthoryear{Chen et~al.}{2015}]{11}
	\begin{barticle}
		\bauthor{\bsnm{Chen}, \binits{X.}},
		\bauthor{\bsnm{Li}, \binits{M.}},
		\bauthor{\bsnm{Yang}, \binits{X.}}:
		\batitle{Stacked denoise autoencoder based feature extraction and
			classification for hyperspectral images}.
		\bjtitle{Journal of Sensors,2016,(2015-11-30)}
		\bvolume{2016},
		\bfpage{1}--\blpage{10}
		(\byear{2015})
	\end{barticle}
	\endbibitem
	
	\bibitem[\protect\citeauthoryear{Mei et~al.}{2019}]{12}
	\begin{barticle}
		\bauthor{\bsnm{Mei}, \binits{S.}},
		\bauthor{\bsnm{Ji}, \binits{J.}},
		\bauthor{\bsnm{Geng}, \binits{Y.}},
		\bauthor{\bsnm{Zhang}, \binits{Z.}},
		\bauthor{\bsnm{Li}, \binits{X.}},
		\bauthor{\bsnm{Du}, \binits{Q.}}:
		\batitle{Unsupervised spatial--spectral feature learning by 3d convolutional
			autoencoder for hyperspectral classification}.
		\bjtitle{IEEE Transactions on Geoscience and Remote Sensing}
		\bvolume{57}(\bissue{9}),
		\bfpage{6808}--\blpage{6820}
		(\byear{2019})
	\end{barticle}
	\endbibitem
	
	\bibitem[\protect\citeauthoryear{Lee and Kwon}{2017}]{13}
	\begin{barticle}
		\bauthor{\bsnm{Lee}, \binits{H.}},
		\bauthor{\bsnm{Kwon}, \binits{H.}}:
		\batitle{Going deeper with contextual cnn for hyperspectral image
			classification}.
		\bjtitle{IEEE Trans Image Process}
		\bvolume{26}(\bissue{10}),
		\bfpage{4843}--\blpage{4855}
		(\byear{2017})
	\end{barticle}
	\endbibitem
	
	\bibitem[\protect\citeauthoryear{Zhang et~al.}{2017}]{14}
	\begin{barticle}
		\bauthor{\bsnm{Zhang}, \binits{H.}},
		\bauthor{\bsnm{Li}, \binits{Y.}},
		\bauthor{\bsnm{Zhang}, \binits{Y.}},
		\bauthor{\bsnm{Shen}, \binits{Q.}}:
		\batitle{Spectral-spatial classification of hyperspectral imagery using a
			dual-channel convolutional neural network}.
		\bjtitle{Remote Sensing Letters}
		\bvolume{8}(\bissue{4-6}),
		\bfpage{438}--\blpage{447}
		(\byear{2017})
	\end{barticle}
	\endbibitem
	
	\bibitem[\protect\citeauthoryear{Zhong et~al.}{2017}]{15}
	\begin{barticle}
		\bauthor{\bsnm{Zhong}, \binits{Z.}},
		\bauthor{\bsnm{Li}, \binits{J.}},
		\bauthor{\bsnm{Luo}, \binits{Z.}},
		\bauthor{\bsnm{Chapman}, \binits{M.}}:
		\batitle{Spectral--spatial residual network for hyperspectral image
			classification: A 3-d deep learning framework}.
		\bjtitle{IEEE Transactions on Geoscience and Remote Sensing}
		\bvolume{56}(\bissue{2}),
		\bfpage{847}--\blpage{858}
		(\byear{2017})
	\end{barticle}
	\endbibitem
	
	\bibitem[\protect\citeauthoryear{Liu et~al.}{2018}]{16}
	\begin{barticle}
		\bauthor{\bsnm{Liu}, \binits{B.}},
		\bauthor{\bsnm{Yu}, \binits{X.}},
		\bauthor{\bsnm{Zhang}, \binits{P.}},
		\bauthor{\bsnm{Tan}, \binits{X.}},
		\bauthor{\bsnm{Wang}, \binits{R.}},
		\bauthor{\bsnm{Zhi}, \binits{L.}}:
		\batitle{Spectral--spatial classification of hyperspectral image using
			three-dimensional convolution network}.
		\bjtitle{Journal of Applied Remote Sensing}
		\bvolume{12}(\bissue{1}),
		\bfpage{016005}--\blpage{016005}
		(\byear{2018})
	\end{barticle}
	\endbibitem
	
	\bibitem[\protect\citeauthoryear{Mou et~al.}{2017}]{17}
	\begin{barticle}
		\bauthor{\bsnm{Mou}, \binits{L.}},
		\bauthor{\bsnm{Ghamisi}, \binits{P.}},
		\bauthor{\bsnm{Zhu}, \binits{X.X.}}:
		\batitle{Deep recurrent neural networks for hyperspectral image
			classification}.
		\bjtitle{IEEE Transactions on Geoscience and Remote Sensing}
		\bvolume{55}(\bissue{7}),
		\bfpage{3639}--\blpage{3655}
		(\byear{2017})
	\end{barticle}
	\endbibitem
	
	\bibitem[\protect\citeauthoryear{Liu et~al.}{2018}]{18}
	\begin{barticle}
		\bauthor{\bsnm{Liu}, \binits{B.}},
		\bauthor{\bsnm{Yu}, \binits{X.}},
		\bauthor{\bsnm{Yu}, \binits{A.}},
		\bauthor{\bsnm{Zhang}, \binits{P.}},
		\bauthor{\bsnm{Wan}, \binits{G.}}:
		\batitle{Spectral-spatial classification of hyperspectral imagery based on
			recurrent neural networks}.
		\bjtitle{Remote sensing letters}
		\bvolume{9}(\bissue{10-12}),
		\bfpage{1118}--\blpage{1127}
		(\byear{2018})
	\end{barticle}
	\endbibitem
	
	\bibitem[\protect\citeauthoryear{Zhou et~al.}{2019}]{19}
	\begin{barticle}
		\bauthor{\bsnm{Zhou}, \binits{F.}},
		\bauthor{\bsnm{Hang}, \binits{R.}},
		\bauthor{\bsnm{Liu}, \binits{Q.}},
		\bauthor{\bsnm{Yuan}, \binits{X.}}:
		\batitle{Hyperspectral image classification using spectral-spatial lstms -
			sciencedirect}.
		\bjtitle{Neurocomputing}
		\bvolume{328},
		\bfpage{39}--\blpage{47}
		(\byear{2019})
	\end{barticle}
	\endbibitem
	
	\bibitem[\protect\citeauthoryear{Qin et~al.}{2018}]{20}
	\begin{barticle}
		\bauthor{\bsnm{Qin}, \binits{A.}},
		\bauthor{\bsnm{Shang}, \binits{Z.}},
		\bauthor{\bsnm{Tian}, \binits{J.}},
		\bauthor{\bsnm{Wang}, \binits{Y.}},
		\bauthor{\bsnm{Zhang}, \binits{T.}},
		\bauthor{\bsnm{Tang}, \binits{Y.Y.}}:
		\batitle{Spectral--spatial graph convolutional networks for semisupervised
			hyperspectral image classification}.
		\bjtitle{IEEE Geoscience and Remote Sensing Letters}
		\bvolume{16}(\bissue{2}),
		\bfpage{241}--\blpage{245}
		(\byear{2018})
	\end{barticle}
	\endbibitem
	
	\bibitem[\protect\citeauthoryear{Wan et~al.}{2020}]{21}
	\begin{barticle}
		\bauthor{\bsnm{Wan}, \binits{S.}},
		\bauthor{\bsnm{Gong}, \binits{C.}},
		\bauthor{\bsnm{Zhong}, \binits{P.}},
		\bauthor{\bsnm{Pan}, \binits{S.}},
		\bauthor{\bsnm{Li}, \binits{G.}},
		\bauthor{\bsnm{Yang}, \binits{J.}}:
		\batitle{Hyperspectral image classification with context-aware dynamic graph
			convolutional network}.
		\bjtitle{IEEE Transactions on Geoscience and Remote Sensing}
		\bvolume{59}(\bissue{1}),
		\bfpage{597}--\blpage{612}
		(\byear{2020})
	\end{barticle}
	\endbibitem
	
	\bibitem[\protect\citeauthoryear{Hong et~al.}{2020}]{22}
	\begin{barticle}
		\bauthor{\bsnm{Hong}, \binits{D.}},
		\bauthor{\bsnm{Gao}, \binits{L.}},
		\bauthor{\bsnm{Yao}, \binits{J.}},
		\bauthor{\bsnm{Zhang}, \binits{B.}},
		\bauthor{\bsnm{Plaza}, \binits{A.}},
		\bauthor{\bsnm{Chanussot}, \binits{J.}}:
		\batitle{Graph convolutional networks for hyperspectral image classification}.
		\bjtitle{IEEE Transactions on Geoscience and Remote Sensing}
		\bvolume{59}(\bissue{7}),
		\bfpage{5966}--\blpage{5978}
		(\byear{2020})
	\end{barticle}
	\endbibitem
	
	\bibitem[\protect\citeauthoryear{Tang et~al.}{2019}]{23}
	\begin{barticle}
		\bauthor{\bsnm{Tang}, \binits{H.}},
		\bauthor{\bsnm{Li}, \binits{Y.}},
		\bauthor{\bsnm{Han}, \binits{X.}},
		\bauthor{\bsnm{Huang}, \binits{Q.}},
		\bauthor{\bsnm{Xie}, \binits{W.}}:
		\batitle{A spatial--spectral prototypical network for hyperspectral remote
			sensing image}.
		\bjtitle{IEEE Geoscience and Remote Sensing Letters}
		\bvolume{17}(\bissue{1}),
		\bfpage{167}--\blpage{171}
		(\byear{2019})
	\end{barticle}
	\endbibitem
	
	\bibitem[\protect\citeauthoryear{Ma et~al.}{2019}]{24}
	\begin{barticle}
		\bauthor{\bsnm{Ma}, \binits{X.}},
		\bauthor{\bsnm{Ji}, \binits{S.}},
		\bauthor{\bsnm{Wang}, \binits{J.}},
		\bauthor{\bsnm{Geng}, \binits{J.}},
		\bauthor{\bsnm{Wang}, \binits{H.}}:
		\batitle{Hyperspectral image classification based on two-phase relation
			learning network}.
		\bjtitle{IEEE Transactions on Geoscience and Remote Sensing}
		\bvolume{57}(\bissue{12}),
		\bfpage{10398}--\blpage{10409}
		(\byear{2019})
	\end{barticle}
	\endbibitem
	
	\bibitem[\protect\citeauthoryear{Li et~al.}{2021}]{25}
	\begin{barticle}
		\bauthor{\bsnm{Li}, \binits{Z.}},
		\bauthor{\bsnm{Liu}, \binits{M.}},
		\bauthor{\bsnm{Chen}, \binits{Y.}},
		\bauthor{\bsnm{Xu}, \binits{Y.}},
		\bauthor{\bsnm{Du}, \binits{Q.}}:
		\batitle{Deep cross-domain few-shot learning for hyperspectral image
			classification}.
		\bjtitle{IEEE Transactions on Geoscience and Remote Sensing}
		\bvolume{PP}(\bissue{99}),
		\bfpage{1}--\blpage{18}
		(\byear{2021})
	\end{barticle}
	\endbibitem
	
	\bibitem[\protect\citeauthoryear{Bai et~al.}{2022}]{26}
	\begin{barticle}
		\bauthor{\bsnm{Bai}, \binits{J.}},
		\bauthor{\bsnm{Huang}, \binits{S.}},
		\bauthor{\bsnm{Xiao}, \binits{Z.}},
		\bauthor{\bsnm{Li}, \binits{X.}},
		\bauthor{\bsnm{Zhu}, \binits{Y.}},
		\bauthor{\bsnm{Regan}, \binits{A.C.}},
		\bauthor{\bsnm{Jiao}, \binits{L.}}:
		\batitle{Few-shot hyperspectral image classification based on adaptive
			subspaces and feature transformation}.
		\bjtitle{IEEE Transactions on Geoscience and Remote Sensing}
		\bvolume{60},
		\bfpage{1}--\blpage{17}
		(\byear{2022})
	\end{barticle}
	\endbibitem
	
	\bibitem[\protect\citeauthoryear{Zhang et~al.}{2022}]{48}
	\begin{botherref}
		\oauthor{\bsnm{Zhang}, \binits{Y.}},
		\oauthor{\bsnm{Li}, \binits{W.}},
		\oauthor{\bsnm{Zhang}, \binits{M.}},
		\oauthor{\bsnm{Wang}, \binits{S.}},
		\oauthor{\bsnm{Tao}, \binits{R.}},
		\oauthor{\bsnm{Du}, \binits{Q.}}:
		Graph information aggregation cross-domain few-shot learning for hyperspectral
		image classification.
		IEEE Transactions on Neural Networks and Learning Systems
		(2022)
	\end{botherref}
	\endbibitem
	
	\bibitem[\protect\citeauthoryear{Khosla et~al.}{2020}]{54}
	\begin{barticle}
		\bauthor{\bsnm{Khosla}, \binits{P.}},
		\bauthor{\bsnm{Teterwak}, \binits{P.}},
		\bauthor{\bsnm{Wang}, \binits{C.}},
		\bauthor{\bsnm{Sarna}, \binits{A.}},
		\bauthor{\bsnm{Tian}, \binits{Y.}},
		\bauthor{\bsnm{Isola}, \binits{P.}},
		\bauthor{\bsnm{Maschinot}, \binits{A.}},
		\bauthor{\bsnm{Liu}, \binits{C.}},
		\bauthor{\bsnm{Krishnan}, \binits{D.}}:
		\batitle{Supervised contrastive learning}.
		\bjtitle{Advances in neural information processing systems}
		\bvolume{33},
		\bfpage{18661}--\blpage{18673}
		(\byear{2020})
	\end{barticle}
	\endbibitem

}

	\bibitem[\protect\citeauthoryear{Koch et~al.}{2015}]{27}
	\begin{botherref}
		\oauthor{\bsnm{Koch}, \binits{G.}},
		\oauthor{\bsnm{Zemel}, \binits{R.}},
		\oauthor{\bsnm{Salakhutdinov}, \binits{R.}}, et al.:
		Siamese neural networks for one-shot image recognition.
		ICML deep learning workshop
		\textbf{2}(1)
		(2015)
	\end{botherref}
	\endbibitem


\color{black}{
	\bibitem[\protect\citeauthoryear{Vinyals et~al.}{2016}]{28}
	\begin{botherref}
		\oauthor{\bsnm{Vinyals}, \binits{O.}},
		\oauthor{\bsnm{Blundell}, \binits{C.}},
		\oauthor{\bsnm{Lillicrap}, \binits{T.}},
		\oauthor{\bsnm{Wierstra}, \binits{D.}}, et al.:
		Matching networks for one shot learning.
		Advances in neural information processing systems
		\textbf{29}
		(2016)
	\end{botherref}
	\endbibitem
	
	\bibitem[\protect\citeauthoryear{Snell et~al.}{2017}]{29}
	\begin{botherref}
		\oauthor{\bsnm{Snell}, \binits{J.}},
		\oauthor{\bsnm{Swersky}, \binits{K.}},
		\oauthor{\bsnm{Zemel}, \binits{R.}}:
		Prototypical networks for few-shot learning.
		Advances in neural information processing systems
		\textbf{30}
		(2017)
	\end{botherref}
	\endbibitem
}	

	

	\bibitem[\protect\citeauthoryear{Sung et~al.}{2018}]{30}
	\begin{botherref}
		\oauthor{\bsnm{Sung}, \binits{F.}},
		\oauthor{\bsnm{Yang}, \binits{Y.}},
		\oauthor{\bsnm{Zhang}, \binits{L.}},
		\oauthor{\bsnm{Xiang}, \binits{T.}},
		\oauthor{\bsnm{Torr}, \binits{P.H.}},
		\oauthor{\bsnm{Hospedales}, \binits{T.M.}}:
		Learning to compare: Relation network for few-shot learning.
		Proceedings of the IEEE conference on computer vision and pattern recognition,
		1199--1208
		(2018)
	\end{botherref}
	\endbibitem
	
	\bibitem[\protect\citeauthoryear{Zhang et~al.}{2021}]{31}
	\begin{botherref}
		\oauthor{\bsnm{Zhang}, \binits{B.}},
		\oauthor{\bsnm{Li}, \binits{X.}},
		\oauthor{\bsnm{Ye}, \binits{Y.}},
		\oauthor{\bsnm{Huang}, \binits{Z.}},
		\oauthor{\bsnm{Zhang}, \binits{L.}}:
		Prototype completion with primitive knowledge for few-shot learning.
		Proceedings of the IEEE/CVF Conference on Computer Vision and Pattern
		Recognition,
		3754--3762
		(2021)
	\end{botherref}
	\endbibitem
	
	\bibitem[\protect\citeauthoryear{Xue and Wang}{2020}]{32}
	\begin{barticle}
		\bauthor{\bsnm{Xue}, \binits{W.}},
		\bauthor{\bsnm{Wang}, \binits{W.}}:
		\batitle{One-shot image classification by learning to restore prototypes}.
		\bjtitle{Proceedings of the AAAI Conference on Artificial Intelligence}
		\bvolume{34}(\bissue{04}),
		\bfpage{6558}--\blpage{6565}
		(\byear{2020})
	\end{barticle}
	\endbibitem
	
	\bibitem[\protect\citeauthoryear{Xu et~al.}{2021}]{33}
	\begin{botherref}
		\oauthor{\bsnm{Xu}, \binits{W.}},
		\oauthor{\bsnm{Xu}, \binits{Y.}},
		\oauthor{\bsnm{Wang}, \binits{H.}},
		\oauthor{\bsnm{Tu}, \binits{Z.}}:
		Attentional constellation nets for few-shot learning.
		International Conference on Learning Representations
		(2021)
	\end{botherref}
	\endbibitem
	
	\bibitem[\protect\citeauthoryear{Finn et~al.}{2017}]{34}
	\begin{botherref}
		\oauthor{\bsnm{Finn}, \binits{C.}},
		\oauthor{\bsnm{Abbeel}, \binits{P.}},
		\oauthor{\bsnm{Levine}, \binits{S.}}:
		Model-agnostic meta-learning for fast adaptation of deep networks.
		International conference on machine learning,
		1126--1135
		(2017)
	\end{botherref}
	\endbibitem
	
	\bibitem[\protect\citeauthoryear{Jamal and Qi}{2019}]{35}
	\begin{botherref}
		\oauthor{\bsnm{Jamal}, \binits{M.A.}},
		\oauthor{\bsnm{Qi}, \binits{G.-J.}}:
		Task agnostic meta-learning for few-shot learning.
		Proceedings of the IEEE/CVF Conference on Computer Vision and Pattern
		Recognition,
		11719--11727
		(2019)
	\end{botherref}
	\endbibitem
	
	\bibitem[\protect\citeauthoryear{Garcia and Bruna}{2018}]{36}
	\begin{botherref}
		\oauthor{\bsnm{Garcia}, \binits{V.}},
		\oauthor{\bsnm{Bruna}, \binits{J.}}:
		Few-shot learning with graph neural networks.
		6th International Conference on Learning Representations, ICLR 2018
		(2018)
	\end{botherref}
	\endbibitem
	
	\bibitem[\protect\citeauthoryear{Kim et~al.}{2019}]{37}
	\begin{botherref}
		\oauthor{\bsnm{Kim}, \binits{J.}},
		\oauthor{\bsnm{Kim}, \binits{T.}},
		\oauthor{\bsnm{Kim}, \binits{S.}},
		\oauthor{\bsnm{Yoo}, \binits{C.D.}}:
		Edge-labeling graph neural network for few-shot learning.
		Proceedings of the IEEE/CVF conference on computer vision and pattern
		recognition,
		11--20
		(2019)
	\end{botherref}
	\endbibitem
	
	\bibitem[\protect\citeauthoryear{Chen et~al.}{2021}]{38}
	\begin{botherref}
		\oauthor{\bsnm{Chen}, \binits{C.}},
		\oauthor{\bsnm{Yang}, \binits{X.}},
		\oauthor{\bsnm{Xu}, \binits{C.}},
		\oauthor{\bsnm{Huang}, \binits{X.}},
		\oauthor{\bsnm{Ma}, \binits{Z.}}:
		Eckpn: Explicit class knowledge propagation network for transductive few-shot
		learning.
		Proceedings of the IEEE/CVF Conference on Computer Vision and Pattern
		Recognition,
		6596--6605
		(2021)
	\end{botherref}
	\endbibitem
	
	\bibitem[\protect\citeauthoryear{Zhang et~al.}{2019}]{39}
	\begin{botherref}
		\oauthor{\bsnm{Zhang}, \binits{C.}},
		\oauthor{\bsnm{Lyu}, \binits{X.}},
		\oauthor{\bsnm{Tang}, \binits{Z.}}:
		Tgg: Transferable graph generation for zero-shot and few-shot learning.
		Proceedings of the 27th ACM International Conference on Multimedia,
		1641--1649
		(2019)
	\end{botherref}
	\endbibitem
	
	\bibitem[\protect\citeauthoryear{Ma et~al.}{2020}]{40}
	\begin{barticle}
		\bauthor{\bsnm{Ma}, \binits{Y.}},
		\bauthor{\bsnm{Bai}, \binits{S.}},
		\bauthor{\bsnm{An}, \binits{S.}},
		\bauthor{\bsnm{Liu}, \binits{W.}},
		\bauthor{\bsnm{Liu}, \binits{A.}},
		\bauthor{\bsnm{Zhen}, \binits{X.}},
		\bauthor{\bsnm{Liu}, \binits{X.}}:
		\batitle{Transductive relation-propagation network for few-shot learning.}
		\bjtitle{IJCAI}
		\bvolume{20},
		\bfpage{804}--\blpage{810}
		(\byear{2020})
	\end{barticle}
	\endbibitem
	

\color{black}{
	\bibitem[\protect\citeauthoryear{Wang et~al.}{2018}]{41}
	\begin{barticle}
		\bauthor{\bsnm{Wang}, \binits{W.}},
		\bauthor{\bsnm{Dou}, \binits{S.}},
		\bauthor{\bsnm{Jiang}, \binits{Z.}},
		\bauthor{\bsnm{Sun}, \binits{L.}}:
		\batitle{A fast dense spectral--spatial convolution network framework for
			hyperspectral images classification}.
		\bjtitle{Remote sensing}
		\bvolume{10}(\bissue{7}),
		\bfpage{1068}
		(\byear{2018})
	\end{barticle}
	\endbibitem
	
	\bibitem[\protect\citeauthoryear{Mou et~al.}{2020}]{63}
	\begin{barticle}
		\bauthor{\bsnm{Mou}, \binits{L.}},
		\bauthor{\bsnm{Lu}, \binits{X.}},
		\bauthor{\bsnm{Li}, \binits{X.}},
		\bauthor{\bsnm{Zhu}, \binits{X.X.}}:
		\batitle{Nonlocal graph convolutional networks for hyperspectral image
			classification}.
		\bjtitle{IEEE Transactions on Geoscience and Remote Sensing}
		\bvolume{58}(\bissue{12}),
		\bfpage{8246}--\blpage{8257}
		(\byear{2020})
	\end{barticle}
	\endbibitem
	
	\bibitem[\protect\citeauthoryear{Liu et~al.}{2017}]{42}
	\begin{barticle}
		\bauthor{\bsnm{Liu}, \binits{B.}},
		\bauthor{\bsnm{Yu}, \binits{X.}},
		\bauthor{\bsnm{Zhang}, \binits{P.}},
		\bauthor{\bsnm{Yu}, \binits{A.}},
		\bauthor{\bsnm{Fu}, \binits{Q.}},
		\bauthor{\bsnm{Wei}, \binits{X.}}:
		\batitle{Supervised deep feature extraction for hyperspectral image
			classification}.
		\bjtitle{IEEE Transactions on Geoscience and Remote Sensing}
		\bvolume{56}(\bissue{4}),
		\bfpage{1909}--\blpage{1921}
		(\byear{2017})
	\end{barticle}
	\endbibitem
	
	\bibitem[\protect\citeauthoryear{Huang and Chen}{2020}]{43}
	\begin{barticle}
		\bauthor{\bsnm{Huang}, \binits{L.}},
		\bauthor{\bsnm{Chen}, \binits{Y.}}:
		\batitle{Dual-path siamese cnn for hyperspectral image classification with
			limited training samples}.
		\bjtitle{IEEE Geoscience and Remote Sensing Letters}
		\bvolume{18}(\bissue{3}),
		\bfpage{518}--\blpage{522}
		(\byear{2020})
	\end{barticle}
	\endbibitem
	
	\bibitem[\protect\citeauthoryear{Rao et~al.}{2020}]{44}
	\begin{barticle}
		\bauthor{\bsnm{Rao}, \binits{M.}},
		\bauthor{\bsnm{Tang}, \binits{P.}},
		\bauthor{\bsnm{Zhang}, \binits{Z.}}:
		\batitle{A developed siamese cnn with 3d adaptive spatial-spectral pyramid
			pooling for hyperspectral image classification}.
		\bjtitle{Remote Sensing}
		\bvolume{12}(\bissue{12}),
		\bfpage{1964}
		(\byear{2020})
	\end{barticle}
	\endbibitem
	
	\bibitem[\protect\citeauthoryear{Sun et~al.}{2022}]{45}
	\begin{barticle}
		\bauthor{\bsnm{Sun}, \binits{J.}},
		\bauthor{\bsnm{Shen}, \binits{X.}},
		\bauthor{\bsnm{Sun}, \binits{Q.}}:
		\batitle{Hyperspectral image few-shot classification network based on the earth
			mover's distance}.
		\bjtitle{IEEE Transactions on Geoscience and Remote Sensing}
		\bvolume{60},
		\bfpage{1}--\blpage{14}
		(\byear{2022})
	\end{barticle}
	\endbibitem
	
	\bibitem[\protect\citeauthoryear{Gao et~al.}{2020}]{46}
	\begin{barticle}
		\bauthor{\bsnm{Gao}, \binits{K.}},
		\bauthor{\bsnm{Liu}, \binits{B.}},
		\bauthor{\bsnm{Yu}, \binits{X.}},
		\bauthor{\bsnm{Qin}, \binits{J.}},
		\bauthor{\bsnm{Zhang}, \binits{P.}},
		\bauthor{\bsnm{Tan}, \binits{X.}}:
		\batitle{Deep relation network for hyperspectral image few-shot
			classification}.
		\bjtitle{Remote Sensing}
		\bvolume{12}(\bissue{6}),
		\bfpage{923}
		(\byear{2020})
	\end{barticle}
	\endbibitem
	
	\bibitem[\protect\citeauthoryear{Rao et~al.}{2019}]{47}
	\begin{barticle}
		\bauthor{\bsnm{Rao}, \binits{M.}},
		\bauthor{\bsnm{Tang}, \binits{P.}},
		\bauthor{\bsnm{Zhang}, \binits{Z.}}:
		\batitle{Spatial--spectral relation network for hyperspectral image
			classification with limited training samples}.
		\bjtitle{IEEE Journal of Selected Topics in Applied Earth Observations and
			Remote Sensing}
		\bvolume{12}(\bissue{12}),
		\bfpage{5086}--\blpage{5100}
		(\byear{2019})
	\end{barticle}
	\endbibitem
	
	\bibitem[\protect\citeauthoryear{Zuo et~al.}{2022}]{62}
	\begin{barticle}
		\bauthor{\bsnm{Zuo}, \binits{X.}},
		\bauthor{\bsnm{Yu}, \binits{X.}},
		\bauthor{\bsnm{Liu}, \binits{B.}},
		\bauthor{\bsnm{Zhang}, \binits{P.}},
		\bauthor{\bsnm{Tan}, \binits{X.}}:
		\batitle{Fsl-egnn: Edge-labeling graph neural network for hyperspectral image
			few-shot classification}.
		\bjtitle{IEEE Transactions on Geoscience and Remote Sensing}
		\bvolume{60},
		\bfpage{1}--\blpage{18}
		(\byear{2022})
	\end{barticle}
	\endbibitem
	
	\bibitem[\protect\citeauthoryear{Xi et~al.}{2022}]{61}
	\begin{barticle}
		\bauthor{\bsnm{Xi}, \binits{B.}},
		\bauthor{\bsnm{Li}, \binits{J.}},
		\bauthor{\bsnm{Li}, \binits{Y.}},
		\bauthor{\bsnm{Song}, \binits{R.}},
		\bauthor{\bsnm{Hong}, \binits{D.}},
		\bauthor{\bsnm{Chanussot}, \binits{J.}}:
		\batitle{Few-shot learning with class-covariance metric for hyperspectral image
			classification}.
		\bjtitle{IEEE Transactions on Image Processing}
		\bvolume{31},
		\bfpage{5079}--\blpage{5092}
		(\byear{2022})
	\end{barticle}
	\endbibitem
	
	\bibitem[\protect\citeauthoryear{Vaswani et~al.}{2017}]{55}
	\begin{botherref}
		\oauthor{\bsnm{Vaswani}, \binits{A.}},
		\oauthor{\bsnm{Shazeer}, \binits{N.}},
		\oauthor{\bsnm{Parmar}, \binits{N.}},
		\oauthor{\bsnm{Uszkoreit}, \binits{J.}},
		\oauthor{\bsnm{Jones}, \binits{L.}},
		\oauthor{\bsnm{Gomez}, \binits{A.N.}},
		\oauthor{\bsnm{Kaiser}, \binits{{\L}.}},
		\oauthor{\bsnm{Polosukhin}, \binits{I.}}:
		Attention is all you need.
		Advances in neural information processing systems
		\textbf{30}
		(2017)
	\end{botherref}
	\endbibitem
	
	\bibitem[\protect\citeauthoryear{Long et~al.}{2018}]{56}
	\begin{botherref}
		\oauthor{\bsnm{Long}, \binits{M.}},
		\oauthor{\bsnm{Cao}, \binits{Z.}},
		\oauthor{\bsnm{Wang}, \binits{J.}},
		\oauthor{\bsnm{Jordan}, \binits{M.I.}}:
		Conditional adversarial domain adaptation.
		Advances in neural information processing systems
		\textbf{31}
		(2018)
	\end{botherref}
	\endbibitem
	
	\bibitem[\protect\citeauthoryear{Li et~al.}{2017}]{57}
	\begin{barticle}
		\bauthor{\bsnm{Li}, \binits{Y.}},
		\bauthor{\bsnm{Zhang}, \binits{H.}},
		\bauthor{\bsnm{Shen}, \binits{Q.}}:
		\batitle{Spectral--spatial classification of hyperspectral imagery with 3d
			convolutional neural network}.
		\bjtitle{Remote Sensing}
		\bvolume{9}(\bissue{1}),
		\bfpage{67}
		(\byear{2017})
	\end{barticle}
	\endbibitem

}

	\bibitem[\protect\citeauthoryear{Yu et~al.}{2019}]{58}
	\begin{bchapter}
		\bauthor{\bsnm{Yu}, \binits{C.}},
		\bauthor{\bsnm{Wang}, \binits{J.}},
		\bauthor{\bsnm{Chen}, \binits{Y.}},
		\bauthor{\bsnm{Huang}, \binits{M.}}:
		\bctitle{Transfer learning with dynamic adversarial adaptation network}.
		\bjtitle{IEEE Transactions on Geoscience and Remote Sensing}
		pp. \bfpage{778}--\blpage{786}
		(\byear{2019})
	\end{bchapter}
	\endbibitem


\color{black}{

	\bibitem[\protect\citeauthoryear{Zhu et~al.}{2020}]{59}
	\begin{barticle}
		\bauthor{\bsnm{Zhu}, \binits{Y.}},
		\bauthor{\bsnm{Zhuang}, \binits{F.}},
		\bauthor{\bsnm{Wang}, \binits{J.}},
		\bauthor{\bsnm{Ke}, \binits{G.}},
		\bauthor{\bsnm{Chen}, \binits{J.}},
		\bauthor{\bsnm{Bian}, \binits{J.}},
		\bauthor{\bsnm{Xiong}, \binits{H.}},
		\bauthor{\bsnm{He}, \binits{Q.}}:
		\batitle{Deep subdomain adaptation network for image classification}.
		\bjtitle{IEEE transactions on neural networks and learning systems}
		\bvolume{32}(\bissue{4}),
		\bfpage{1713}--\blpage{1722}
		(\byear{2020})
	\end{barticle}
	\endbibitem
	
	\bibitem[\protect\citeauthoryear{Liu et~al.}{2018}]{60}
	\begin{barticle}
		\bauthor{\bsnm{Liu}, \binits{B.}},
		\bauthor{\bsnm{Yu}, \binits{X.}},
		\bauthor{\bsnm{Yu}, \binits{A.}},
		\bauthor{\bsnm{Zhang}, \binits{P.}},
		\bauthor{\bsnm{Wan}, \binits{G.}},
		\bauthor{\bsnm{Wang}, \binits{R.}}:
		\batitle{Deep few-shot learning for hyperspectral image classification}.
		\bjtitle{IEEE Transactions on Geoscience and Remote Sensing}
		\bvolume{57}(\bissue{4}),
		\bfpage{2290}--\blpage{2304}
		(\byear{2018})
	\end{barticle}
	\endbibitem
}
\end{thebibliography}

\end{document}